%% file: main.tex
\documentclass[10pt,twocolumn,letterpaper]{article}

\usepackage{cvpr}              % To produce the CAMERA-READY version

\usepackage{algorithm}
\usepackage{algorithmic}
\usepackage{float}
\usepackage{booktabs}
\usepackage{array}
\usepackage{multirow}
\usepackage{utfsym}
\usepackage{amssymb}
\usepackage{amsmath}
\usepackage{amsfonts}
\usepackage{rotating}
\usepackage[accsupp]{axessibility} % Improves PDF readability for those with disabilities.


\definecolor{cvprblue}{rgb}{0.21,0.49,0.74}
\usepackage[pagebackref,breaklinks,colorlinks,allcolors=cvprblue]{hyperref}

\def\paperID{10676} % *** Enter the Paper ID here
\def\confName{CVPR}
\def\confYear{2025}

\title{SCSegamba: Lightweight Structure-Aware Vision Mamba for Crack Segmentation in Structures}

\author{
  Hui Liu\textsuperscript{1,2,3}, 
  Chen Jia\textsuperscript{1,2,3,*}\thanks{*Corresponding author.}, 
  Fan Shi\textsuperscript{1,2,3},
  Xu Cheng\textsuperscript{1,2,3},
  Shengyong Chen\textsuperscript{1,2,3}
  \\
  \textsuperscript{1}School of Computer Science and Engineering, Tianjin University of Technology \\
  \textsuperscript{2}Engineering Research Center of Learning-Based Intelligent System (Ministry of Education) \\
  \textsuperscript{3}Key Laboratory of Computer Vision and System (Ministry of Education) \\
  {\tt\small 
  liuhui1109@stud.tjut.edu.cn, 
  \{jiachen, shifan\}@email.tjut.edu.cn,
  \{xu.cheng, sy\}@ieee.org
  }
}

\makeatletter
\def\thanks#1{\protected@xdef\@thanks{\@thanks
        \protect\footnotetext{#1}}}
\makeatother

\begin{document}
\maketitle
\input{sec/0_abstract}    
\input{sec/1_intro}

\input{sec/2_relatedworks}

\input{sec/3_methodology}

\input{sec/4_experiments}
\input{sec/5_Conclusion}
\input{sec/6_Acknowledgement}
\input{sec/X_suppl}

% \clearpageCVPR
\clearpage

{
    \small
    \bibliographystyle{ieeenat_fullname}
    \bibliography{main}
}


\end{document}

%% file: sec/0_abstract.tex
\begin{abstract}

\hypertarget{figurea}{Pixel-level} \hypertarget{figureb}{segmentation} \hypertarget{figurec}{of} structural cracks across various scenarios remains a considerable challenge. Current methods encounter challenges in effectively modeling crack morphology and texture, facing challenges in balancing segmentation quality with low computational resource usage. To overcome these limitations, we propose a lightweight Structure-Aware Vision Mamba Network (\textbf{SCSegamba}), capable of generating high-quality pixel-level segmentation maps by leveraging both the morphological information and texture cues of crack pixels with minimal computational cost. Specifically, we developed a \textbf{S}tructure-\textbf{A}ware \textbf{V}isual \textbf{S}tate \textbf{S}pace module (\textbf{SAVSS}), which incorporates a lightweight \textbf{G}ated \textbf{B}ottleneck \textbf{C}onvolution (\textbf{GBC}) and a \textbf{S}tructure-\textbf{A}ware \textbf{S}canning \textbf{S}trategy (\textbf{SASS}). The key insight of GBC lies in its effectiveness in modeling the morphological information of cracks, while the SASS enhances the perception of crack topology and texture by strengthening the continuity of semantic information between crack pixels. Experiments on crack benchmark datasets demonstrate that our method outperforms other state-of-the-art (SOTA) methods, achieving the highest performance with only \textbf{2.8M} parameters. On the multi-scenario dataset, our method reached \textbf{0.8390} in F1 score and \textbf{0.8479} in mIoU. The code is available at \url{https://github.com/Karl1109/SCSegamba}.

% Pixel-level segmentation of structural cracks across various scenarios remains a considerable challenge. Current methods encounter challenges in effectively modeling crack morphology and texture, facing challenges in balancing segmentation quality with low computational resource usage. To overcome these limitations, we propose a lightweight Structure-Aware Vision Mamba Network (SCSegamba), capable of generating high-quality pixel-level segmentation maps by leveraging both the morphological information and texture cues of crack pixels with minimal computational cost. Specifically, we developed a Structure-Aware Visual State Space module (SAVSS), which incorporates a lightweight Gated Bottleneck Convolution (GBC) and a Structure-Aware Scanning Strategy (SASS). The key insight of GBC lies in its effectiveness in modeling the morphological information of cracks, while the SASS enhances the perception of crack topology and texture by strengthening the continuity of semantic information between crack pixels. Experiments on crack benchmark datasets demonstrate that our method outperforms other state-of-the-art (SOTA) methods, achieving the highest performance with only 2.8M parameters. On the multi-scenario dataset, our method reached 0.8390 in F1 score and 0.8479 in mIoU.

\end{abstract}

%% file: sec/1_intro.tex
\vspace{-0.5cm}
\section{Introduction}
\label{sec:intro}

\begin{figure}[htbp]
  \centering
  \includegraphics[width=0.482\textwidth]{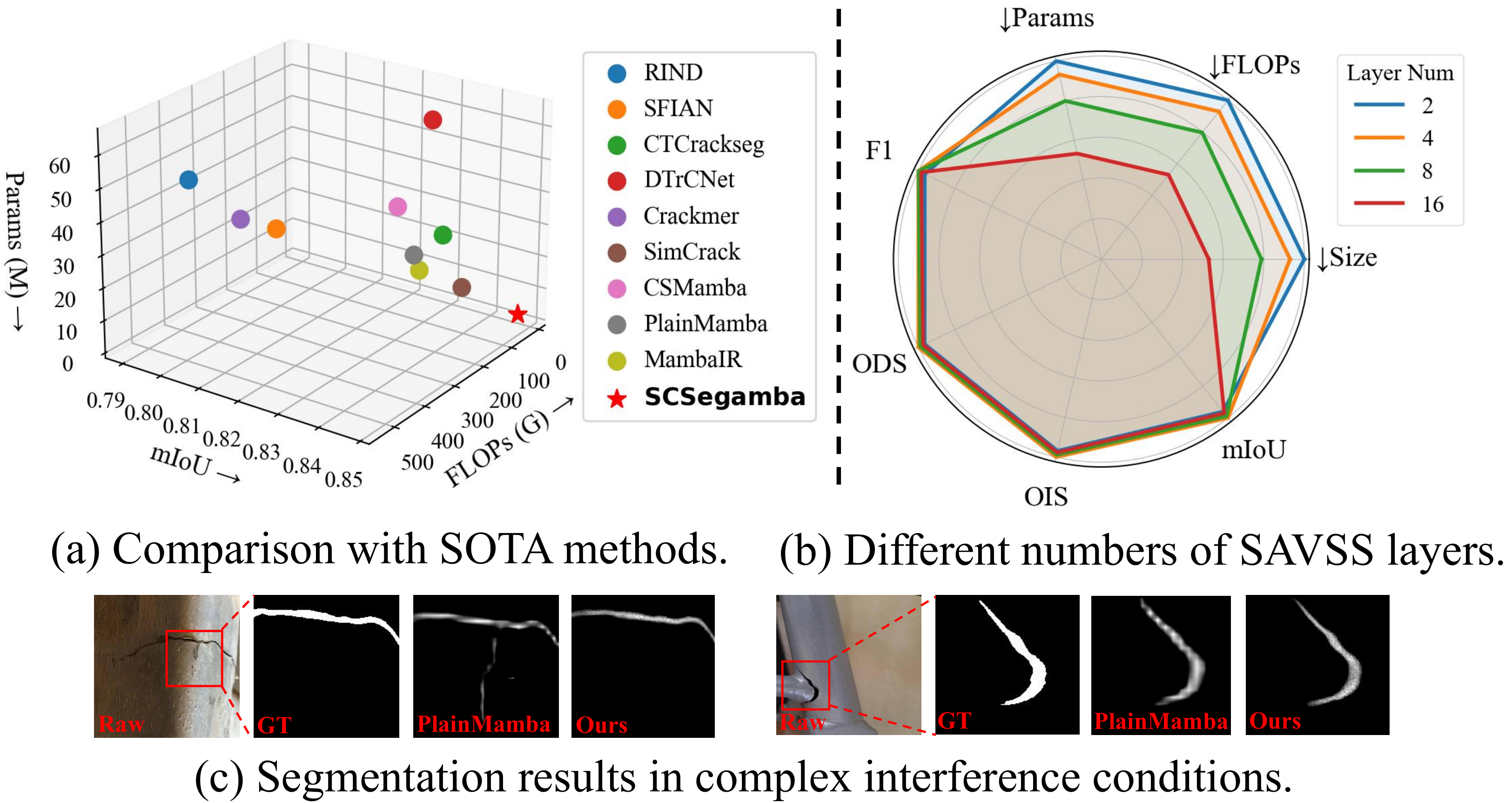}
  \caption{Performance of SCSegamba on multi-scenario TUT \cite{liu2024staircase} dataset. (a) Comparison with SOTA methods. (b) Impact of different SAVSS layer numbers on performance, with normalized metrics; FLOPs (G), Params (M), and Size (MB) decrease towards the edges. (c) Visual results under complex interference.}
  \label{fig:intro}
  \vspace{-0.5cm}
\end{figure}

Structures like bitumen pavement, concrete, and metal frequently develop cracks under shear stress, making regular health monitoring essential to avoid production issues \cite{chen2024mind, chen2022geometry, hsieh2020machine, kheradmandi2022critical, liao2022automatic}. Due to differences in material properties and environmental conditions, various materials exhibit significant variations in crack morphology and visual appearance \cite{lang2024augmented}. As a result, achieving pixel-level crack segmentation across diverse scenarios remains a complex challenge. Recently, Convolutional Neural Networks (CNNs), such as ECSNet \cite{zhang2023ecsnet} and SFIAN \cite{cheng2023selective}, have shown effective crack feature extraction capabilities in segmentation tasks due to their strong local inductive properties. However, their limited receptive field constrains their ability to model broad-scope irregular dependencies across the entire image, resulting in discontinuous segmentation and weak background noise suppression. Although dilated convolution \cite{chen2018encoder} expand the receptive field, their inherent inductive bias still prevents them from fully addressing this issue \cite{zhang2024robust}, especially in complex crack patterns with heavy background interference.

% The success of Vision Transformers (ViT) \cite{alexey2020image, xia2024vit} has demonstrated the effectiveness of Transformers \cite{vaswani2017attention} with self-attention in the visual domain. Transformers are adept at capturing irregular pixel dependencies, which are essential for recognizing complex crack textures. Networks such as SwinCrack \cite{wang2024swincrack}, MFAFNet \cite{dong2024mfafnet}, and CATransUNet \cite{chu2024transformer} achieve high-quality crack segmentation by combining shape cues with these dependencies. 

The success of Vision Transformer (ViT) \cite{alexey2020image, xia2024vit, vaswani2017attention} has demonstrated Transformer's effectiveness in capturing irregular pixel dependencies, which is crucial for recognizing complex crack textures, as seen in networks like DTrCNet \cite{xiang2023crack}, MFAFNet \cite{dong2024mfafnet}, and Crackmer \cite{wang2024dual}. However, the quadratic complexity of attention calculations with sequence length leads to high memory use and training challenges for high-resolution images, limiting deployment on resource-constrained edge devices and practical applications. As shown in Figure \hyperlink{figurea}{1(a)}, Transformer-based methods like CTCrackseg \cite{tao2023convolutional} and DTrCNet \cite{xiang2023crack} perform well, but their large parameter counts and high computational demands limit their deployability on resource-constrained devices. Although variants like Sparse Transformer \cite{child2019generating} and Linear Transformer \cite{katharopoulos2020transformers} reduce computational requirements by sparsifying or linearizing attention, they sacrifice the ability to model irregular dependencies and pixel textures, hindering pixel-level detection tasks. Consequently, Transformer-based methods struggle to balance segmentation performance with computational efficiency.

\begin{figure*}[htbp]
  \centering
  \includegraphics[width=\textwidth]{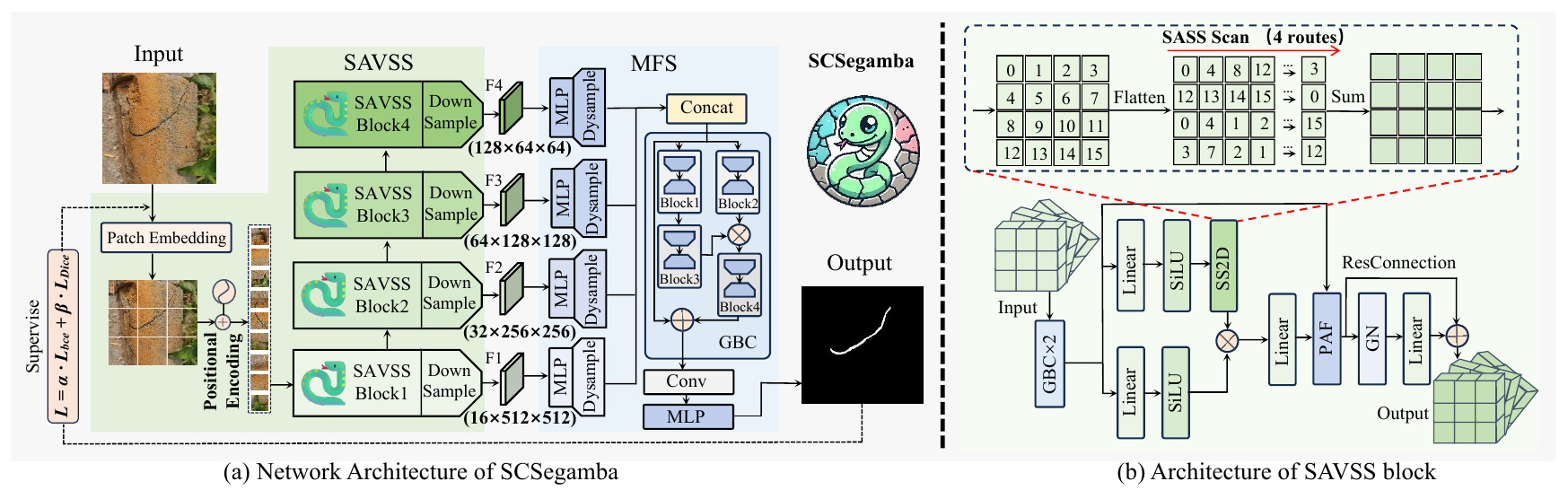}
  \caption{Overview of our proposed method. (a) illustrates the overall architecture of SCSegamba and the processing flow for crack images. (b) displays the structure of the SAVSS block. The input crack image undergoes comprehensive morphological and texture feature extraction through SAVSS, while MFS produces a high-quality pixel-level segmentation map.}
  \label{fig:SCSegamba}
  \vspace{-0.5cm}
\end{figure*}

% The existing VSS block structure and scanning strategies in Mamba are inadequate for effectively capturing both crack morphology and texture. For example, directly processing feature maps through linear layers prevents selective enhancement or suppression of crack region features, limiting the extraction of detailed morphology and texture. Additionally, basic parallel or unidirectional diagonal scanning strategies are suboptimal for capturing the complex topology of cracks, reducing the model's effectiveness in handling regions with high complexity and noise. Consequently, current Mamba-based state-space models often miss or misdetect cracks in multi-scenario, high-noise, and small-crack images. 

Recently, Selective State Space Models (SSMs) have attracted considerable interest due to Mamba's showing strong performance in sequence modeling while maintaining low computational demands \cite{gu2022train, gu2021efficiently}. Vision Mamba (ViM) \cite{vim} and VMamba \cite{liu2024vmamba} have extended Mamba to the visual domain. The irregular extension of crack regions with numerous branches in low-contrast images, often affected by irrelevant areas and shadows, challenges existing Mamba VSS (Visual State Space Model) block structures and scanning strategies in capturing crack morphology and texture effectively. Most Mamba-based methods \cite{liu2024cmunet, yang2024plainmamba,  xing2024segmamba, xie2024fusionmamba} process feature maps through linear layers, limiting selective enhancement or suppression of crack features against irrelevant disturbances, thus reducing detailed morphological extraction. Additionally, common parallel  or unidirectional diagonal scans \cite{guo2024mambair} struggle to maintain semantic continuity when handling irregular, multi-directional pixel topologies, weakening their ability to manage complex textures and suppress noise. Consequently, current Mamba-based SSM frequently produce false or missed detections in multi-scenario crack images. Moreover, although these methods have the advantage of fewer parameters, there remains potential to further reduce their computational demands and enhance their deployability on edge devices. As shown in Figure \hyperlink{figureb}{1(b)}, CSMamba \cite{liu2024cmunet}, MambaIR \cite{guo2024mambair}, and PlainMamba \cite{yang2024plainmamba} showed unsatisfactory performance on crack images, with room for improvement in parameter count and computational load. 

To tackle the challenge of balancing high segmentation quality with low computational demands, we propose the SCSegamba network that produces high-quality pixel-level crack segmentation maps with low computational resources. To improve the Mamba network's perception of irregular crack textures, we design a Structure-Aware Visual State Space block (SAVSS), employing a Structure-Aware Scanning Strategy (SASS) to enhance semantic continuity and strengthen crack morphology perception. For capturing crack shape cues while maintaining low parameter and computational costs, we designed a lightweight Gated Bottleneck Convolution (GBC) that dynamically adjusts weights for complex backgrounds and varying morphologies. Additionally, the Multi-scale Feature Segmentation head (MFS) integrates the GBC and Multi-layer Perceptron (MLP) to achieve high-quality segmentation maps with low computational requirements. As shown in Figure \hyperlink{figurec}{1(c)}, optimal segmentation performance was achieved with four SAVSS layers, producing clear segmentation maps that effectively mask complex interference while maintaining model lightweightness.

In essence, the primary contributions of our work are outlined as follows:

\begin{itemize}

\item We propose a novel lightweight vision Mamba network, SCSegamba, for crack segmentation. This model effectively captures morphological and irregular texture cues of crack pixels, using low computational resources to generate high-quality segmentation maps.

\item We design the SAVSS with a lightweight GBC Convolution and a SASS scanning strategy to enhance the handling and perception of irregular texture cues in crack images. Additionally, a simple yet effective MFS is developed to generate segmentation maps with relatively low computational resources.

% The GBC more effectively captures fine crack structures, while the SASS improves perception of crack topology and contextual continuity.

\item We evaluate SCSegamba on benchmark datasets across diverse scenarios, with results demonstrating that our method outperforms other SOTA methods while maintaining the low parameter count.

\end{itemize}

%% file: sec/2_relatedworks.tex
\vspace{-0.2cm}
\section{Related Works}
\label{sec:relatedworks}

\vspace{-0.1cm}
\subsection{Crack Segmentation Network}
Early crack detection methods often relied on traditional feature extraction techniques, such as wavelet transform \cite{zhou2006wavelet}, percolation models \cite{yamaguchi2008image}, and the k-means algorithm \cite{lattanzi2014robust}. Although straightforward to implement, these methods face challenges in suppressing background interference and achieving high segmentation accuracy. With advancements in deep learning, researchers have developed CNN-based crack segmentation networks that achieve SOTA performance \cite{xiao2018weighted, choi2019sddnet, lei2024joint, qu2021crack, cheng2023selective, he2016deep}. For instance, DeepCrack \cite{liu2019deepcrack} enables end-to-end pixel-level segmentation, while FPHBN \cite{yang2019feature} demonstrates strong generalization capabilities. BARNet \cite{guo2021barnet} detects crack boundaries by integrating image gradients with features, and ADDUNet \cite{al2023asymmetric} captures both fine and coarse crack characteristics across varied conditions. Although CNN-based methods show significant promise, the local operations and limited receptive fields of CNNs limit their ability to fully capture crack texture cues and effectively suppress background noise.

Transformers \cite{vaswani2017attention}, with their self-attention mechanism, are well-suited for visual tasks that require long-range dependency modeling, making them increasingly popular in crack segmentation networks \cite{quan2023crackvit, xu2024boosting, dong2024mfafnet, Liu_2021_ICCV, Liu2023CrackFormer}. For example, VCVNet \cite{qi2024vision}, based on Vision Transformer (ViT), is designed for bridge crack segmentation, addressing fine-grained segmentation challenges. SWT-CNN \cite{li2024hybrid} combines Swin Transformer and CNN for automatic feature extraction, while TBUNet \cite{zhang2024distilling}, a Transformer-based knowledge distillation model, achieves high-performance crack segmentation with a hybrid loss function. Although Transformer-based methods are highly effective at capturing crack texture cues and suppressing background noise, their self-attention mechanism introduces computational complexity that grows quadratically with sequence length. This results in a high parameter count and significant computational demands, which limit their deployment on resource-constrained edge devices.
\vspace{-0.2cm}

\begin{figure}[htbp]
  \centering
  \includegraphics[width=0.45\textwidth]{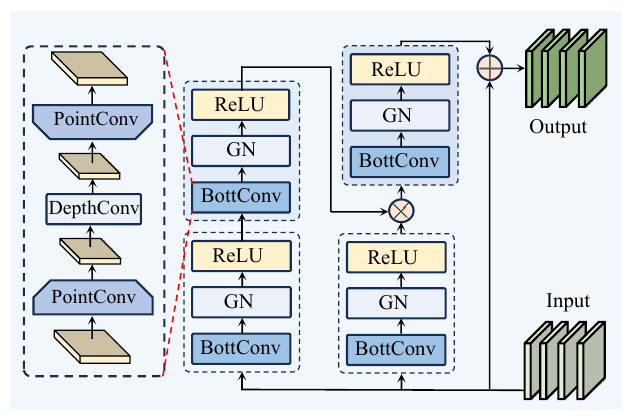}
  \caption{Architecture of GBC. It employs bottleneck convolution to efficiently reduce the parameters and computational load, while the gating mechanism enhances the model's adaptability in processing diverse crack patterns and complex backgrounds. GN represents group normalization.}
  \label{fig:GBC}
    \vspace{-0.6cm}
\end{figure}

\subsection{Selective State Space Model}
The introduction of the Selective State Space Models (S6) in the Mamba model \cite{gu2021efficiently} has highlighted the potential of SSM \cite{goel2022s, gu2022parameterization}. Unlike the linear time-invariant S4 model, S6 efficiently captures complex long-distance dependencies while preserving computational efficiency, achieving strong performance in NLP, audio, and genomics. Consequently, researchers have adapted Mamba to the visual domain, creating various VSS blocks. ViM \cite{vim} achieves comparable modeling to ViT \cite{alexey2020image} without attention mechanisms, using fewer computational resources, while VMamba \cite{liu2024vmamba} prioritizes efficient computation and high performance. PlainMamba \cite{yang2024plainmamba} employs a fixed-width layer stacking approach, excelling in tasks such as instance segmentation and object detection. However, the VSS block and scanning strategy require specific optimizations for each visual task, as tasks differ in their reliance on long- and short-distance information, necessitating customized VSS block designs to ensure optimal performance.

Currently, no high-performing Mamba-based model exists for crack segmentation. Thus, designing an optimized VSS structure specifically for crack segmentation is essential to improve performance and efficiency. Given the intricate details and irregular textures of cracks, the VSS block requires strong shape extraction and directional awareness to effectively capture crack texture cues. Additionally, it should facilitate efficient crack segmentation while minimizing computational resource requirements.

%% file: sec/3_methodology.tex
\vspace{-0.1cm}
\section{Methodology}
\label{sec:methodology}

\subsection{Preliminary}

The complete architecture of our proposed SCSegamba is depicted in Figure
\ref{fig:SCSegamba}. It includes two main components: the SAVSS for extracting crack shape and texture cues, and the MFS for efficient feature processing. To capture key crack region cues, we integrate the GBC at the initial stage of SAVSS and the final stage of MFS.

For a single RGB image \( E \in \mathbb{R}^{ 3 \times H \times W} \), spatial information is divided into \( n \) patches, forming a sequence \( \{B_1, B_2, \dots, B_n\} \). This sequence is processed through the SAVSS block, embedding key crack pixel cues into multi-scale feature maps \( \{F_1, F_2, F_3, F_4\} \). Finally, in the MFS, all information is consolidated into a single tensor, producing a refined segmentation output \( W \in \mathbb{R}^{ 1 \times H \times W } \).

\begin{figure*}[htbp]
  \centering
  \includegraphics[width=0.9\textwidth]{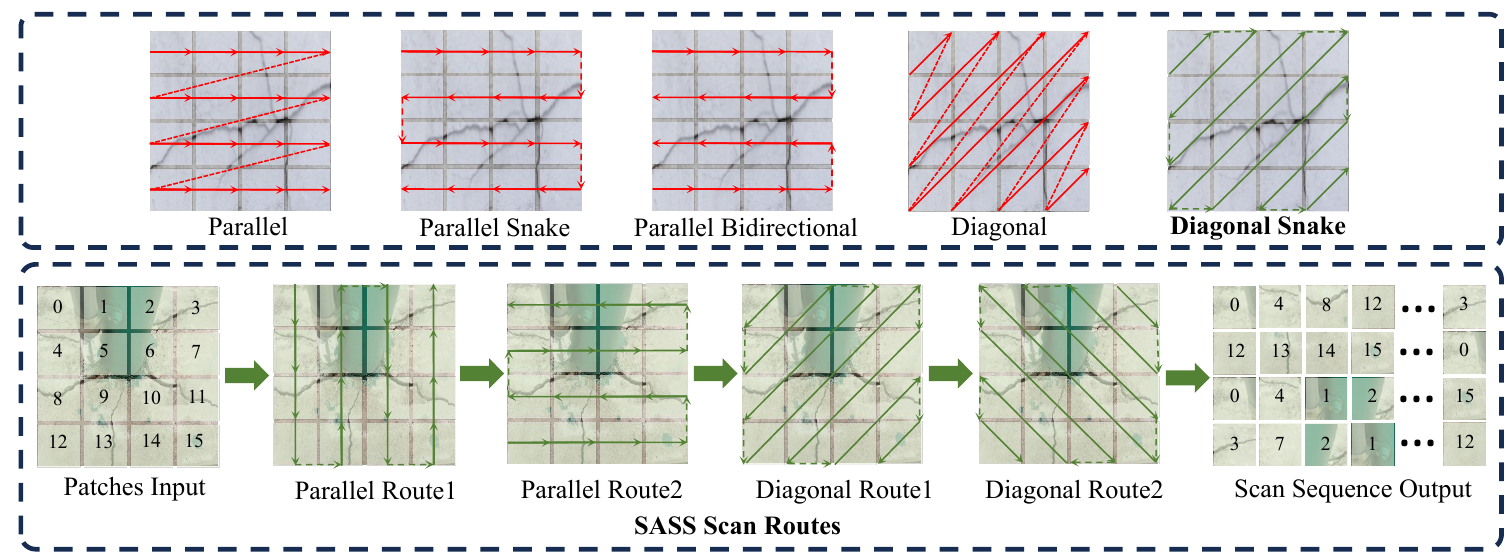}
  \caption{Illustration of our proposed SASS and other scanning strategies. The first row presents four commonly used single scanning paths, along with our proposed diagonal snake path. The second row illustrates the execution flow of our proposed SASS scanning strategy.}
  \label{fig:Scan}
  \vspace{-0.4cm}
\end{figure*}

\subsection{Lightweight Gated Bottleneck Convolution}

The gating mechanism enables dynamic features for each spatial position and channel, enhancing the model's ability to capture details \cite{dauphin2017language, yu2019free}. To further reduce parameter count and computational cost, we embedded a bottleneck convolution (BottConv) with low-rank approximation \cite{li2024lors}, mapping matrices from high to low dimensional spaces and significantly lowering computational complexity.

\begin{figure}[htbp]
  \centering
  \includegraphics[width=0.48\textwidth]{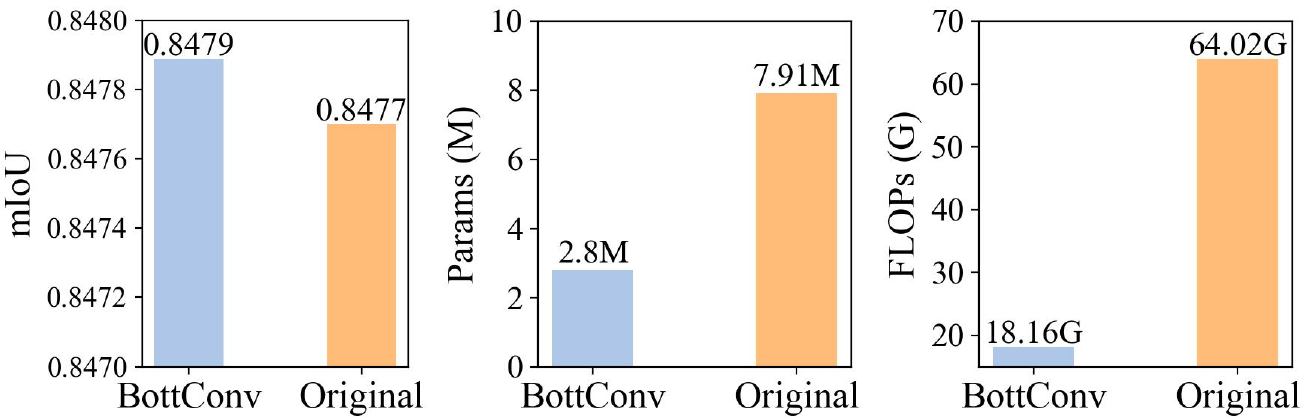}
  \caption{Performance comparison between using BottConv and raw convolution in GBC on the TUT \cite{liu2024staircase} dataset.}
  \label{fig:bott}
  \vspace{-0.6cm}
\end{figure}

In the convolution layer, assuming the spatial size of the filter is \( p \), the number of input channels is \( d \) and the input is \(s\), the convolution response can be represented as:
\vspace{-0.2cm}
\begin{equation}
  z = Qs + c
\end{equation}
where \( Q \) is a matrix of size \( f \times (p^2 \times d) \), \( f \) is the number of output channels, and \( c \) is the original bias term. Assuming \( z \) lies in a low-rank subspace of rank \( f_0 \), it can be represented as \( z = V(z - z_1) + z_1 \), where \( z_1 \) abstracts the mean vector of features, acting as an auxiliary variable to facilitate theoretical derivation and correct feature offsets, \( V = LM^T \) (\( L \in \mathbb{R}^{f \times f_0} \), \( M \in \mathbb{R}^{(p^2d) \times f_0} \)) represents the low-rank projection matrix. The simplified response then becomes:
\vspace{-0.1cm}
\begin{equation} 
  z = LM^Ts + c'
\end{equation}

Since \( f_0 < f \), the computational complexity reduces from \( O(fp^2d) \) to \( O(f_0p^2d) + O(ff_0) \), where \( O(ff_0) \ll O(fp^2d) \), indicating that the computational complexity reduction is proportional to the original ratio of \( f_0/f \).

In BottConv, pointwise convolutions project features into or out of low-rank subspace, thus significantly reducing complexity, while depthwise convolution that perform spatial information-adequate extraction in subspace guarantees negligibly low complexity. As shown in Figure \ref{fig:bott}, BottConv in our GBC design significantly reduces parameter count and computational load compared to the original convolution, with minimal performance impact.

As shown in Figure \ref{fig:GBC}, the input feature \( x \in \mathbb{R}^{ C \times H \times W } \) is retained as \( x_{\text{residual}} = x \) to facilitate the residual connection. Subsequently, the feature \( x \) is passed through the BottConv layer, followed by normalization and activation functions, resulting in the features \( x_1 \) and \( g_2(x) \) as shown below:

\vspace{-0.4cm}
\begin{equation}
  g_1(x) = ReLU(Norm_1(f_1(x)))
\end{equation}
\vspace{-0.4cm}
\begin{equation}
  x_1 = ReLU(Norm_2(BottConv_2(g_1(x))))
\end{equation}
\vspace{-0.4cm}
\begin{equation}
  g_2(x) = ReLU(Norm_3(BottConv_3(x)))
\end{equation}

To generate the gating feature map, \( x_1 \) and \( g_2(x) \) are combined through the Hadamard product:
\vspace{-0.2cm}
\begin{equation}
  m(x) = x_1 \odot g_2(x)
\end{equation}

The gating feature map \( m(x) \) is subsequently processed through BottConv once again to further refine fine-grained details. After the residual connection is applied, the resulting output is:

\vspace{-0.3cm}
\begin{equation}
  y = ReLU(Norm_4(BottConv_4(m(x))))
\end{equation}
\vspace{-0.4cm}
\begin{equation}
  Output = y + x_{residual}
\end{equation}

The design of BottConv and deeper gated branch enable the model to preserve basic crack features while dynamically refining the fine-grained feature characterization of the main branch, resulting in more accurate segmentation maps in detailed regions.

\subsection{Structure-Aware Visual State Space Module}
\label{subsec:savss}

Our designed SAVSS features a two-dimensional selective scan (SS2D) tailored for visual tasks. Different scanning strategies impact the model's ability to capture continuous crack textures. As shown in Figure \ref{fig:Scan}, current vision Mamba networks use various scanning directions, including parallel, snake, bidirectional, and diagonal scans \cite{vim, liu2024vmamba, yang2024plainmamba}. Parallel and diagonal scans lack continuity across rows or diagonals, which limits their sensitivity to crack directions. Although bidirectional and snake scans maintain semantic continuity along horizontal or vertical paths, they struggle to capture diagonal or interwoven textures. To address this, our proposed diagonal snake scanning is designed to better capture complex crack texture cues.

SASS consists of four paths: two parallel snake paths and two diagonal snake paths. This design enables the effective extraction of continuous semantic information in regular crack regions while preserving texture continuity in multiple directions, making it suitable for multi-scenario crack images with complex backgrounds.

After the RGB crack image undergoes Patch Embedding and Position Encoding, it is input as a sequence into the SAVSS block. To maintain a lightweight network, we use only 4 layers of SAVSS blocks. The processing equations are as follows:

% \vspace{-0.4cm}
% \begin{equation}
%   \overline{A} = e^{\Delta A}
% \end{equation}
% \begin{equation}
%   \overline{B} = (\Delta A)^{-1} (e^{\Delta A} - I) \cdot \Delta B
% \end{equation}
% \begin{equation}
%   h_k = \overline{A} h_{k-1} + \overline{B} x_k
% \end{equation}
% \begin{equation}
%   y_k = C h_k + D x_k
% \end{equation}

% In these equations, the input  \( x \in \mathbb{R}^{d \times L} \), \( A \in \mathbb{R}^{H \times L} \) controls the hidden spatial state, \( D \in \mathbb{R}^{L \times L} \) is used to initialize the skip connection for input, \( h_k \) represents the specific hidden state at time step \( k \), and \( B \in \mathbb{R}^{H \times L} \) and \( C \in \mathbb{R}^{H \times L} \) are matrices with hidden spatial dimensions \( H \) and temporal dimensions \( L \), respectively, obtained through selective scanning SS2D. These are trainable parameters that are updated accordingly. \( y_k \) represents the output at time step \( k \).

\vspace{-0.4cm}
\begin{equation}
  \overline{P} = e^{\Delta P}
\end{equation}
\vspace{-0.4cm}
\begin{equation}
  \overline{Q} = (\Delta P)^{-1} (e^{\Delta P} - I) \cdot \Delta Q
\end{equation}
\vspace{-0.4cm}
\begin{equation}
  z_k = \overline{P} z_{k-1} + \overline{Q} w_k
\end{equation}
\vspace{-0.4cm}
\begin{equation}
  u_k = R z_k + S w_k
\end{equation}

In these equations, the input  \( w \in \mathbb{R}^{t \times D} \), \( P \in \mathbb{R}^{G \times D} \) controls the hidden spatial state, \( S \in \mathbb{R}^{D \times D} \) is used to initialize the skip connection for input, \( z_k \) represents the specific hidden state at time step \( k \), and \( Q \in \mathbb{R}^{G \times D} \) and \( R \in \mathbb{R}^{G \times D} \) are matrices with hidden spatial dimensions \( G \) and temporal dimensions \( D \), respectively, obtained through selective scanning SS2D. These are trainable parameters that are updated accordingly. \( u_k \) represents the output at time step \( k \). SASS establishes multi-directional adjacency relationships, allowing the hidden state \( z_k \) to capture more intricate topological and textural details, while enabling the output \( u_k \) to more effectively integrate multi-directional features.

To effectively combine the initial sequence \( x \) with the sequence processed through SS2D, we incorporate Pixel Attention-oriented Fusion (PAF) \cite{liu2024staircase}, enhancing SAVSS’s ability to capture crack shape and texture details. Following selective scanning, a residual connection is applied to the fused information to preserve detail and facilitate feature flow. Furthermore, GBC refines the inter-layer output within SAVSS, strengthening crack information extraction and boosting performance in later stages.

\subsection{Multi-scale Feature Segmentation Head}

% Given the lightweight requirements of the model and the necessity to effectively capture multiscale feature representations, we incorporated a Multi-Layer Perceptron (MLP) into the segmentation head. 
Unlike convolutional layers, the MLP swiftly learns the mapping relationships between features and labels, thereby reducing model complexity. When the four feature maps \( F_1, F_2, F_3, F_4 \in \mathbb{R}^{ C \times H \times W } \) produced by SAVSS are fed into MFS, they undergo individual processing through the efficient MLP operation and dynamic upsampling \cite{liu2023learning}, restoring their resolution to the original size and yielding \( F_1^{\text{up}}, F_2^{\text{up}}, F_3^{\text{up}}, F_4^{\text{up}} \in \mathbb{R}^{ C \times H \times W } \). The formula is as follows:

\vspace{-0.2cm}
\begin{equation}
  F_i^{up} = DySample(MLP_i(F_i))
\end{equation}
where \( i \) denotes the layer index.

To integrate all multi-scale crack shape and texture representations, these feature maps are aggregated into a single tensor, obtaining a high-quality crack segmentation map \( o \in \mathbb{R}^{ 1 \times H \times W } \) as follows:
\begin{equation}
  o_1 = GBC(Concat(F_1^{up}, F_2^{up}, F_3^{up}, F_4^{up}))
\end{equation}
\vspace{-0.4cm}
\begin{equation}
  o = MLP(Conv(o_1))
\end{equation}

\subsection{Objective Function}
We use a blend of Binary Cross-Entropy loss (BCE) \cite{li2024rediscovering} and Dice loss \cite{sudre2017generalised} as the objective function, which helps improve the network's robustness to imbalanced pixel data. The overall loss function is expressed as follows:

% \begin{equation}
%   L_{Dice} = 1 - \frac{2 \sum_{j=1}^M p_j \hat{p}_j + \epsilon}{\sum_{j=1}^M p_j + \sum_{j=1}^M \hat{p}_j + \epsilon}
% \end{equation}

% \begin{equation}
% L_{BCE} = -E \left[ p_j \log(\hat{p}_j) + (1 - p_j) \log(1 - \hat{p}_j) \right]
% \end{equation}

\vspace{-0.3cm}
\begin{equation}
\label{eq:loss}
  L = \alpha \cdot L_{Dice} + \beta \cdot L_{BCE}
\end{equation}
% where \( M \) denotes the number of samples, \( p_j \) is the ground truth label for the \( j \)-th sample, \( \hat{p}_j \) is the predicted probability for the \( j \)-th sample, \( \epsilon \) is a small constant.
where the hyperparameters \( \alpha \) and \( \beta \) control the weights of the two loss components. The ratio of \( \alpha \) to \( \beta \) is set to 1:5.
\vspace{-0.2cm}

%% file: sec/4_experiments.tex
\section{Experiments}
\label{sec:experiments}
\begin{figure*}[htbp]
  \centering
  \includegraphics[width=\textwidth]{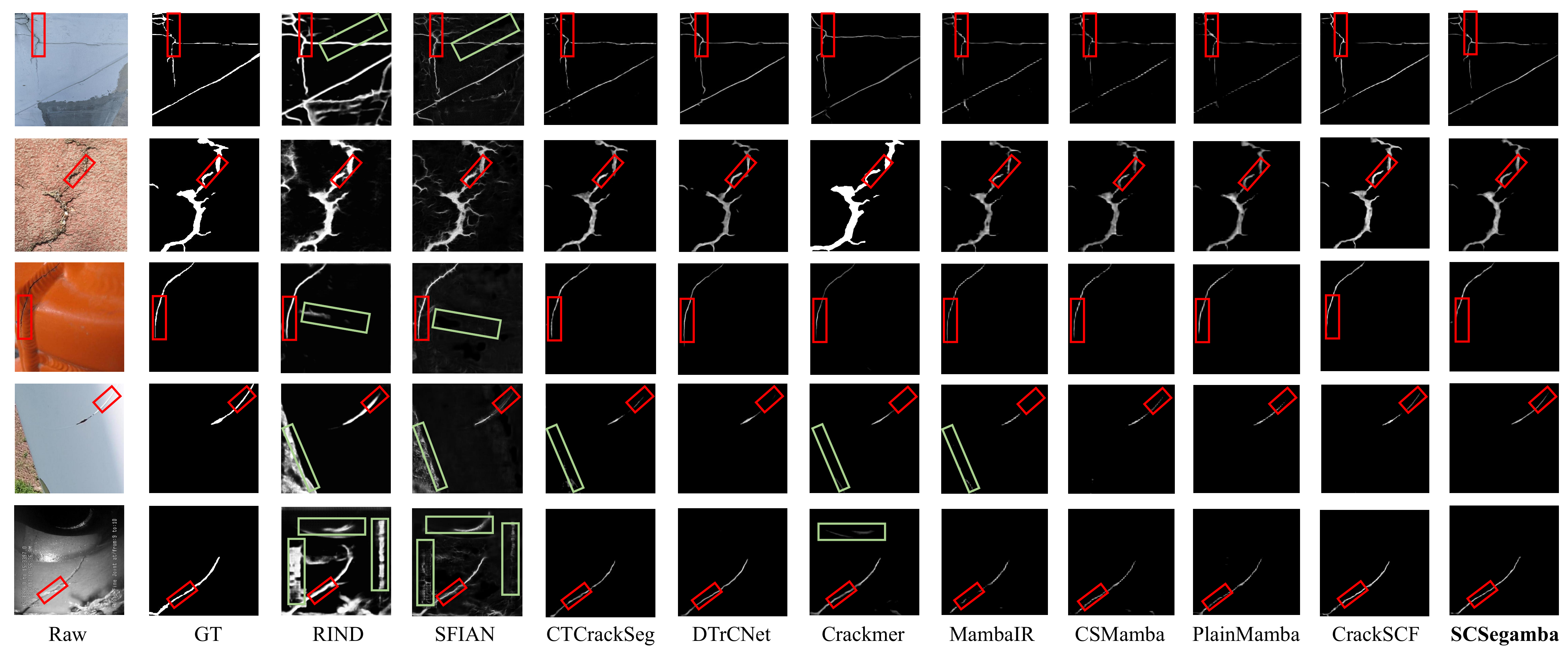}
  \caption{Visual comparison of typical cracks with 9 SOTA methods across four datasets. Red boxes highlight critical details, and green boxes mark misidentified regions.}
  \label{fig:Com_SOTA}
  \vspace{-0.5cm}
\end{figure*}

\subsection{Datasets}

\noindent \textbf{Crack500} \cite{yang2019feature} \textbf{.} The images in this dataset were captured using a mobile phone. The original dataset consists of 500 bitumen crack images, which were expanded to 3368 images through data augmentation. Each image is paired with a corresponding pixel-level annotated binary image.

\noindent \textbf{DeepCrack} \cite{liu2019deepcrack} \textbf{.} The dataset comprises 537 RGB images of cement, bricks and bitumen cracks under various conditions, including fine, wide, stained, and fuzzy cracks, ensuring diversity and representativeness.

\noindent \textbf{CrackMap} \cite{katsamenis2023few} \textbf{.} The dataset was created for a road crack segmentation study and consists of 120 high-resolution RGB images capturing a variety of thin and complex bitumen road cracks.

\noindent \textbf{TUT} \cite{liu2024staircase} \textbf{.} In contrast to other datasets with simple backgrounds, this dataset includes a dense, cluttered background and features cracks with elaborate, intricate shapes. It contains 1408 RGB images across eight scenarios: bitumen, cement, bricks, runway, tiles, metal, blades, and pipes.

During processing, all datasets were divided into training, validation, and test sets with a 7:1:2 ratio.

\subsection{Implementation Details.}
\vspace{-0.13cm}
\noindent \textbf{Experimental Settings.} We built our SCSegamba network using PyTorch v1.13.1 and trained it on an Intel Xeon Platinum 8336C CPU with eight Nvidia GeForce RTX 4090 GPUs. The AdamW optimizer was used with an initial learning rate of 5e-4, PolyLR scheduling, a weight decay of 0.01, and a random seed of 42. The network was trained for 50 epochs, and the model with the best validation performance was selected for testing.

\noindent \textbf{Comparison Methods.} To comprehensively evaluate our model, we compared SCSegamba with 9 SOTA methods. The CNN or Transformer-based models included RIND \cite{pu2021rindnet}, SFIAN \cite{cheng2023selective}, CTCrackSeg \cite{tao2023convolutional}, DTrCNet \cite{xiang2023crack}, Crackmer \cite{wang2024dual} and SimCrack \cite{jaziri2024designing}. Additionally, we compared it with other Mamba-based models, including CSMamba \cite{liu2024cmunet}, PlainMamba \cite{yang2024plainmamba}, and MambaIR \cite{guo2024mambair}.

\noindent \textbf{Evaluation Metrics.} We used six metrics to evaluate SCSegamba’s performance: Precision (P), Recall (R), F1  Score ($F1 = \frac{2RP}{R + P}$), Optimal Dataset Scale (ODS), Optimal Image Scale (OIS), and mean Intersection over Union (mIoU). ODS measures the model's adaptability to datasets of varying scales at a fixed threshold m, while OIS evaluates adaptability across image scales at an optimal threshold n. The calculation formulas are as follows:
\begin{equation}
ODS = \max_m \frac{2 \cdot P_m \cdot R_m}{P_m + R_m}
\end{equation}
\vspace{-0.4cm}
\begin{equation}
OIS = \frac{1}{N} \sum_{i=1}^{N} \max_n \frac{2 \cdot P_{n,i} \cdot R_{n,i}}{P_{n,i} + R_{n,i}}
\end{equation}

mIoU is used to measure the mean proportion of the intersection over union between the ground truth and the predicted results. The calculation is given by the formula:

\vspace{-0.4cm}
\begin{equation}
mIoU = \frac{1}{N+1} \sum_{l=0}^{N} \frac{p_{ll}}{\sum_{t=0}^{N} p_{lt} + \sum_{t=0}^{N} p_{tl} - p_{ll}}
\end{equation}
where \( N \) is the number of classes, which we set as \( N = 1 \); \( t \) represents the ground truth, \( l \) represents the predicted value, and \( p_{tl} \) represents the count of pixels classified as \( l \) but belonging to \( t \).

Additionally, we evaluated our method's complexity using three metrics: FLOPs, Params, and Model Size, representing computational complexity, parameter complexity, and memory footprint.

\begin{table*}[htbp]
    \centering
    \begin{tabular}{l | *{5}{>{\centering\arraybackslash}p{0.75cm}} p{0.83cm} | *{6}{>{\centering\arraybackslash}p{0.77cm}}}
    \toprule 
        \multirow{2}{*}{Methods} & ~ & ~ & Crack500 & ~ & ~ & ~ & ~ & ~ & DeepCrack & ~ & ~ & ~ \\ \cline{2-13} 
        & ODS & OIS & P & R & F1 & mIoU & ODS & OIS & P & R & F1 & mIoU \\ \hline
        RIND \cite{pu2021rindnet} & 0.6469  & 0.6483  & 0.6998  & 0.7245  & 0.7119  & 0.7381  & 0.8087  & 0.8267  & 0.7896  & \underline{0.8920}  & 0.8377  & 0.8391  \\ 
        SFIAN \cite{cheng2023selective} & 0.6977  & \underline{0.7348}  & 0.6983  & 0.7742  & 0.7343  & 0.7604  & 0.8616  & \underline{0.8928}  & 0.8549  & 0.8692  & 0.8620  & 0.8776  \\ 
        CTCrackSeg \cite{tao2023convolutional} & 0.6941  & 0.7059  & 0.6940  & 0.7748  & 0.7322  & 0.7591  & \underline{0.8819}  & 0.8904  & 0.9011  & 0.8895  & 0.8952  & \underline{0.8925}  \\ 
        DTrCNet \cite{xiang2023crack} & 0.7012  & 0.7241  & 0.6527  & \textbf{0.8280}  & 0.7357  & 0.7627  & 0.8473  & 0.8512  & 0.8905  & 0.8251  & 0.8566  & 0.8661  \\ 
        Crackmer \cite{wang2024dual} & 0.6933  & 0.7097  & 0.6985  & 0.7572  & 0.7267  & 0.7591  & 0.8712  & 0.8785  & 0.8946  & 0.8783  & 0.8864  & 0.8844  \\ 
        SimCrack \cite{jaziri2024designing} & \underline{0.7127}  & 0.7308  & 0.7093  & \underline{0.7984}  & \underline{0.7516}  & \underline{0.7715}  & 0.8570  & 0.8722  & 0.8984  & 0.8549  & 0.8761  & 0.8744  \\ 
        CSMamba \cite{liu2024cmunet} & 0.6931  & 0.7162  & 0.6858  & 0.7823  & 0.7315  & 0.7592  & 0.8738  & 0.8766  & 0.9025  & 0.8863  & 0.8943  & 0.8863  \\ 
        PlainMamba \cite{yang2024plainmamba} & 0.7035  & 0.7173  & 0.7170  & 0.7557  & 0.7358  & 0.7682  & 0.8646  & 0.8668  & 0.9050  & 0.8659  & 0.8850  & 0.8788  \\ 
        MambaIR \cite{guo2024mambair} & 0.7043  & 0.7189  & \underline{0.7204}  & 0.7681  & 0.7435  & 0.7663  & 0.8796  & 0.8840  & \underline{0.9056}  & 0.8895  & \underline{0.8975}  & 0.8907  \\ 
        SCSegamba (\textbf{Ours}) & \textbf{0.7244}  & \textbf{0.7370}  & \textbf{0.7270}  & 0.7859  & \textbf{0.7553}  & \textbf{0.7778}  & \textbf{0.8938}  & \textbf{0.8990}  & \textbf{0.9097}  & \textbf{0.9124}  & \textbf{0.9110}  & \textbf{0.9022}  \\ \hline
        \multirow{2}{*}{Methods} & ~ & ~ & CrackMap & ~ & ~ & ~ & ~ & ~ &  
  TUT & ~ & ~ & ~ \\ \cline{2-13} 
        & ODS & OIS & P & R & F1 & mIoU & ODS & OIS & P & R & F1 & mIoU \\ \hline
        RIND \cite{pu2021rindnet} & 0.6745  & 0.6943  & 0.6023  & 0.7586  & 0.6699  & 0.7425  & 0.7531  & 0.7891  & 0.7872  & 0.7665  & 0.7767  & 0.8051  \\ 
        SFIAN \cite{cheng2023selective} & 0.7200  & 0.7465  & 0.6715  & 0.7668  & 0.7160  & 0.7748  & 0.7290  & 0.7513  & 0.7715  & 0.7367  & 0.7537  & 0.7896  \\ 
        CTCrackSeg \cite{tao2023convolutional} & 0.7289  & 0.7373  & 0.6911  & \underline{0.7669}  & 0.7270  & 0.7785  & 0.7940  & 0.7996  & \underline{0.8202}  & 0.8195  & 0.8199  & 0.8301  \\ 
        DTrCNet \cite{xiang2023crack} & 0.7328  & 0.7413  & 0.6912  & 0.7681  & 0.7276  & 0.7812  & \underline{0.7987}  & 0.8073  & 0.7972  & 0.8441  & 0.8202  & 0.8331  \\ 
        Crackmer \cite{wang2024dual} & 0.7395  & 0.7437  & 0.7229  & 0.7467  & 0.7346  & 0.7860  & 0.7429  & 0.7640  & 0.7501  & 0.7656  & 0.7578  & 0.7966  \\ 
        SimCrack \cite{jaziri2024designing} & \underline{0.7559}  & \underline{0.7625}  & 0.7380  & 0.7672  & \underline{0.7523}  & \underline{0.7963}  & 0.7984  & \underline{0.8090}  & 0.8051  & 0.8371  & \underline{0.8208}  & \underline{0.8334}  \\ 
        CSMamba \cite{liu2024cmunet} & 0.7371  & 0.7413  & 0.7053  & 0.7663  & 0.7346  & 0.7841  & 0.7879  & 0.7946  & 0.7947  & 0.8353  & 0.8146  & 0.8263  \\ 
        PlainMamba \cite{yang2024plainmamba} & 0.7150  & 0.7189  & 0.6649  & 0.7616  & 0.7099  & 0.7699  & 0.7867  & 0.7967  & 0.7701  & \underline{0.8523}  & 0.8102  & 0.8253  \\ 
        MambaIR \cite{guo2024mambair} & 0.7332  & 0.7347  & \underline{0.7569}  & 0.7013  & 0.7280  & 0.7834  & 0.7861  & 0.7930  & 0.7877  & 0.8387  & 0.8125  & 0.8249  \\ 
        SCSegamba (\textbf{Ours}) & \textbf{0.7741}  & \textbf{0.7766}  & \textbf{0.7629}  & \textbf{0.7727}  & \textbf{0.7678}  & \textbf{0.8094}  & \textbf{0.8204}  & \textbf{0.8255}  & \textbf{0.8241}  & \textbf{0.8545}  & \textbf{0.8390}  & \textbf{0.8479}  \\ \bottomrule 
    \end{tabular} 
    \caption{Comparison with 9 SOTA methods across 4 datasets. Best results are in bold, and second-best results are underlined.}
    \label{tab:SOTA_Com}
    \vspace{-0.4cm}
\end{table*}

\begin{table}[htbp]
    \centering
    \begin{tabular}{l | p{0.7cm} | >{\centering\arraybackslash}p{0.92cm} >{\centering\arraybackslash}p{0.8cm} >{\centering\arraybackslash}p{0.92cm}}
    \toprule
        Methods & Year & FLOPs↓ & Params↓ & Size↓ \\ \hline
        RIND \cite{pu2021rindnet} & 2021 & 695.77G & 59.39M & 453MB \\ 
        SFIAN \cite{cheng2023selective} & 2023 & 84.57G & 13.63M & 56MB \\ 
        CTCrackSeg \cite{tao2023convolutional} & 2023 & 39.47G & 22.88M & 174MB \\ 
        DTrCNet \cite{xiang2023crack} & 2023 & 123.20G & 63.45M & 317MB \\ 
        Crackmer \cite{wang2024dual} & 2024 & \textbf{14.94G} & \underline{5.90M} & \underline{43MB} \\ 
        SimCrack \cite{jaziri2024designing} & 2024 & 286.62G & 29.58M & 225MB \\ 
        CSMamba \cite{liu2024cmunet} & 2024 & 145.84G & 35.95M & 233MB \\ 
        PlainMamba \cite{yang2024plainmamba} & 2024 & 73.36G & 16.72M & 96MB \\ 
        MambaIR \cite{guo2024mambair} & 2024 & 47.32G & 10.34M & 79MB \\ 
        SCSegamba (\textbf{Ours}) & 2024 & \underline{18.16G} & \textbf{2.80M} & \textbf{37MB} \\
    \bottomrule
    \end{tabular}
    \caption{Comparison of complexity with other methods. Best results are in bold, and second-best results are underlined.}
    \label{tab:Complex_Com}
    \vspace{-0.65cm}
\end{table}

\begin{table*}[!ht]
    \centering
    \begin{tabular}{lccccccccc}
    \toprule 
        Seg Head & ODS & OIS & P & R & F1 & mIoU & Params ↓ & FLOPs ↓ & Model Size ↓ \\ \hline
        UNet \cite{ronneberger2015u} & \underline{0.8055}  & \underline{0.8151}  & 0.8148  & \underline{0.8376}  & \underline{0.8260}  & \underline{0.8378}  & 2.92M & 19.27G & 39MB \\ 
        Ham \cite{geng2021attention} & 0.7703  & 0.7784  & 0.7962  & 0.7838  & 0.7909  & 0.8124  & 2.86M & 35.08G & 38MB \\ 
        SegFormer \cite{xie2021segformer} & 0.7947  & 0.7983  & \underline{0.8170}  & 0.8174  & 0.8172  & 0.8307  & \textbf{2.79M} & \textbf{17.87G} & \textbf{35MB} \\ 
        \textbf{MFS} & \textbf{0.8204}  & \textbf{0.8255}  & \textbf{0.8241}  & \textbf{0.8545}  & \textbf{0.8390}  & \textbf{0.8479}  & \underline{2.80M} & \underline{18.16G} & \underline{37MB} \\ 
    \bottomrule
    \end{tabular}
    \caption{Ablation study of different segmentation heads. UNet \cite{ronneberger2015u}, Ham \cite{geng2021attention}, and SegFormer \cite{xie2021segformer} are high-performance heads.}
    \label{tab:abl_head}
      \vspace{-0.2cm}
\end{table*}

\begin{table*}[!ht]
    \centering
    \begin{tabular}{cccccccccccc}
    \toprule
        GBC & PAF & Res & ODS & OIS & P & R & F1 & mIoU & Params ↓ & FLOPs ↓ & Model Size ↓ \\ \hline
        \usym{2713} & \usym{2717} & \usym{2717} & \underline{0.8136}  & 0.8196  & 0.8213  & 0.8461  & \underline{0.8335}  & \underline{0.8434}  & 2.49M & 16.75G & \underline{34MB} \\ 
        \usym{2717} & \usym{2713} & \usym{2717} & 0.7998  & 0.8084  & 0.7918  & \underline{0.8524}  & 0.8222  & 0.8343  & \textbf{2.28M} & \textbf{14.91G} & \textbf{33MB} \\ 
        \usym{2717} & \usym{2717} & \usym{2713} & 0.7936  & 0.8069  & 0.7952  & 0.8438  & 0.8197  & 0.8313  & \underline{2.48M} & \underline{15.65G} & 35MB \\ 
        \usym{2713} & \usym{2713} & \usym{2717} & 0.8047  & 0.8102  & 0.8174  & 0.8379  & 0.8275  & 0.8377  & 2.54M & 17.08G & 35MB \\ 
        \usym{2713} & \usym{2717} & \usym{2713} & 0.8116  & \underline{0.8200}  & 0.8156  & 0.8522  & 0.8334  & 0.8425  & 2.75M & 17.82G & 37MB \\ 
        \usym{2717} & \usym{2713} & \usym{2713} & 0.8023  & 0.8076  & \underline{0.8219}  & 0.8302  & 0.8260  & 0.8360  & 2.54M & 15.99G & 35MB \\ 
        \usym{2713} & \usym{2713} & \usym{2713} & \textbf{0.8204}  & \textbf{0.8255}  & \textbf{0.8241}  & \textbf{0.8545}  & \textbf{0.8390}  & \textbf{0.8479}  & 2.80M & 18.16G & 37MB \\ 
    \bottomrule
    \end{tabular}
    \caption{Ablation study of components within the SAVSS block. Best results are in bold, and second-best results are underlined.}
    \label{tab:abl_mamba_blcok}
    \vspace{-0.65cm}
\end{table*}

\vspace{-0.1cm}
\subsection{Comparison with SOTA Methods}
\label{subsec:com_sota}
\vspace{-0.1cm}

As listed in Table \ref{tab:SOTA_Com}, compared with 9 other SOTA methods, our proposed SCSegamba achieves the best performance across four public datasets. Specifically, on the Crack500 \cite{yang2019feature} and DeepCrack \cite{liu2019deepcrack} datasets, which contain larger and more complex crack regions, SCSegamba achieved the highest performance. Notably, on the DeepCrack dataset, it surpassed the next best method by 1.50\% in F1 score and 1.09\% in mIoU. This improvement is due to the robust ability of GBC to capture morphological clues in large crack areas, enhancing the model's representational power. On the CrackMap \cite{katsamenis2023few} dataset, which features thinner and more elongated cracks, our method surpasses all other SOTA methods in every metric, outperforming the next best method by 2.06\% in F1 and 1.65\% in mIoU. This demonstrates the effectiveness of SASS in capturing fine textures and elongated crack structures. As illustrated in Figure \ref{fig:Com_SOTA}, our method produces clearer and more precise feature maps, with superior detail capture in typical scenarios such as cement and bitumen, compared to other methods.

For the TUT dataset \cite{liu2024staircase}, which includes eight diverse scenarios, our method achieved the best performance, surpassing the next best method by 2.21\% in F1 and 1.74\% in mIoU. As shown in Figure \ref{fig:Com_SOTA}, whether in the complex crack topology of plastic tracks, the noise-heavy backgrounds of metallic materials and turbine blades, or the low-contrast, dimly lit underground pipeline images, SCSegamba consistently produced high-quality segmentation maps while effectively suppressing irrelevant noise. This demonstrates that our method, with the enhanced crack morphology and texture perception from GBC and SASS, exhibits exceptional robustness and stability. Additionally, leveraging MFS for feature aggregation improves multi-scale perception, making our model particularly suited for diverse, interference-rich scenarios.
\vspace{-0.15cm}

\subsection{Complexity Analysis}
\vspace{-0.15cm}
% Table \ref{tab:Complex_Com} compares our method's complexity with other SOTA methods. Our method leads with only 2.8M Params and 36M Model Size, which are 41.54\% and 16.28\% lower than the next lowest results, respectively. Additionally, our method exceeds Crackmer’s FLOPs by only 3.22G, despite Crackmer’s focus on reducing computational complexity. This demonstrates our method’s ability to achieve high-quality segmentation in complex crack scenarios with minimal parameters and computational requirements.
Table \ref{tab:Complex_Com} shows a comparison of the complexity of our method with other SOTA methods when the input image size is uniformly set to 512. With only 2.80M parameters and a model size of 37MB, our method surpasses all others, being 52.54\% and 13.95\% lower than the next best result, respectively. Additionally, compared to Crackmer \cite{wang2024dual}, which prioritizes computational efficiency, our method’s FLOPs are only 3.22G higher. This demonstrates that the combination of lightweight SAVSS and MFS enables high-quality segmentation in noisy crack scenes with minimal parameters and low computational load, which is essential for resource-constrained devices.

\vspace{-0.2cm}
\subsection{Ablation Studies}
\vspace{-0.15cm}
We performed ablation experiments on the representative multi-scenario dataset TUT \cite{liu2024staircase}.

\noindent \textbf{Ablation study of segmentation heads.} As listed in Table \ref{tab:abl_head}, with our designed MFS, SCSegamba achieved the best results across all six metrics, with F1 and mIoU scores 1.57\% and 1.21\% higher than the second-best method. In terms of complexity, although Params, FLOPs, and Model Size are only 0.01M, 0.29G, and 2MB larger than the SegFormer head, our method surpasses it in F1 and mIoU by 2.67\% and 2.07\%, respectively. This demonstrates that MFS enhances SAVSS output integration, significantly improving performance while keeping the model lightweight.

\begin{table}[!ht]
\vspace{-0.15cm}
    \centering
    \begin{tabular}{p{0.85cm}>
    {\centering\arraybackslash}p{0.75cm} >{\centering\arraybackslash}p{0.75cm} >{\centering\arraybackslash}p{0.75cm} >{\centering\arraybackslash}p{0.75cm} >{\centering\arraybackslash}p{0.75cm} >{\centering\arraybackslash}p{0.75cm}}
    \toprule
        Scan & ODS & OIS & P & R & F1 & mIoU \\ \hline
        Parallel & 0.8123  & 0.8184  & 0.8146  & \underline{0.8523}  & 0.8330  & 0.8427  \\ 
        Diag & 0.8091 & 0.8148 & 0.8225  & 0.8417  & 0.8320  & 0.8410  \\ 
        ParaSna & 0.8102 & 0.8162 & 0.8219  & 0.8365  & 0.8291  & 0.8408  \\ 
        DiagSna & \underline{0.8153} & \underline{0.8215} & \underline{0.8237} & 0.8497 & \underline{0.8365} & \underline{0.8451}  \\ 
        \textbf{SASS} & \textbf{0.8204} & \textbf{0.8255} & \textbf{0.8241} & \textbf{0.8545}  & \textbf{0.8390}  & \textbf{0.8479}  \\
    \bottomrule
    \end{tabular}
    \caption{Ablation studies with different four-route scanning strategies in the SAVSS block, comparing parallel, diagonal, parallel snake, and diagonal snake scanning. The impact of different scanning strategies on complexity is negligible; thus, complexity analysis is omitted from this table. Best results are in bold, and second-best results are underlined.}
    \label{tab:abl_scan}
    \vspace{-0.2cm}
\end{table}

\noindent \textbf{Ablation study of components.} Table \ref{tab:abl_mamba_blcok} shows the impact of each component in SAVSS on model performance. When fully utilizing GBC, PAF, and residual connections, our model achieved the best results across all metrics. Notably, adding GBC led to significant improvements in F1 and mIoU by 1.57\% and 1.42\%, respectively, highlighting its strength in capturing crack morphology cues. Similarly, residual connections boosted F1 and mIoU by 0.13\% and 2.47\%, indicating their role in focusing on essential crack features. Although using only PAF resulted in the lowest Params, FLOPs, and Model Size, it significantly reduced performance. These findings demonstrate that our fully integrated SAVSS effectively captures crack morphology and texture cues, achieving top pixel-level segmentation results while maintaining a lightweight model.

\noindent \textbf{Ablation studies of scanning strategies.} As listed in Table \ref{tab:abl_scan}, under the same conditions of using four different directional scanning paths, the model achieved the best performance with our designed SASS scanning strategy, improving F1 and mIoU by 0.30\% and 0.33\% over the diagonal snake strategy. This demonstrates SASS’s ability to construct semantic and dependency information suited to crack topology, enhancing crack pixel perception in subsequent modules. \textbf{More comprehensive experiments and real-world deployments are available in the Appendix.}
\vspace{-0.3cm}

%% file: sec/5_conclusion.tex
\section{Conclusion}
\label{sec:conclusion}
\vspace{-0.2cm}

In this paper, we proposed SCSegamba, a lightweight structure-aware Vision Mamba for precise pixel-level crack segmentation. SCSegamba combines SAVSS and MFS to enhance crack shape and texture perception with a low parameter count. Equipped with GBC and SASS scanning, SAVSS, captures irregular crack textures across various structures. Experiments on four datasets show SCSegamba’s exceptional performance, especially in complex, noisy scenarios. On the challenging multi-scenario dataset, it achieved an F1 score of 0.8390 and mIoU of 0.8479 with only 18.16G FLOPs and 2.8M parameters, demonstrating its effectiveness for real-world crack detection and suitability for edge devices. Future work will incorporate multimodal cues to enhance segmentation quality, while further optimizing VSS design and scan strategies to achieve high-quality results with low computational resources.

\vspace{-0.3cm}

%% file: sec/6_Acknowledgement.tex
\section{Acknowledgement}
\label{sec:acknowledgement}
\vspace{-0.2cm}
This work was supported by the National Natural Science Foundation of China (NSFC) under Grants 62272342, 62020106004, 62306212, and T2422015; the Tianjin Natural Science Foundation under Grants 23JCJQJC00070 and 24PTLYHZ00320; and the Marie Skłodowska-Curie Actions (MSCA) under Project No. 101111188.

%% file: sec/X_suppl.tex
\clearpage
\setcounter{page}{1}
\maketitlesupplementary

\section{Details of SASS and Ablation Experiments}
\label{sec:SASS_Details}

As described in Subsection \ref{subsec:savss}, the SASS strategy enhances semantic capture in complex crack regions by scanning texture cues from multiple directions. SASS combines parallel snake and diagonal snake scans, aligning the scanning paths with the actual extension and irregular shapes of cracks, ensuring comprehensive capture of texture information.

To evaluate the necessity of using four scanning paths in SASS, we conducted ablation experiments with different path numbers across various scanning strategies on multi-scenario dataset TUT. As listed in Table \ref{tab:scan_num}, all strategies performed significantly better with four paths than with two, likely because four paths allow SAVSS to capture finer crack details and topological cues. Notably, aside from SASS, the diagonal snake-like scan consistently achieved the second-best results, with two-path configurations yielding F1 and mIoU scores 0.48\% and 0.45\% higher than the diagonal unidirectional scan. This indicates that the diagonal snake-like scan provides more continuous semantic information, enhancing segmentation. Importantly, our proposed SASS achieved the best results with both two-path and four-path setups, demonstrating its effectiveness in capturing diverse crack topologies.

To clarify the implementation of our proposed SASS, we present its execution process in Algorithm \ref{alg:sass}.

\begin{table}[htbp]
  \centering
    \begin{tabular}{p{0.04cm} |
    >{\centering\arraybackslash}p{0.2cm} |
    >{\centering\arraybackslash}p{0.7cm} 
    >{\centering\arraybackslash}p{0.7cm} 
    >{\centering\arraybackslash}p{0.7cm} 
    >{\centering\arraybackslash}p{0.7cm} 
    >{\centering\arraybackslash}p{0.7cm} 
    >{\centering\arraybackslash}p{0.75cm}} 
    \toprule
    \multirow{2}[9]{*}{\scriptsize\begin{sideways}Parallel\end{sideways}} & \textbf{N} & \textbf{ODS} & \textbf{OIS} & \textbf{P} & \textbf{R} & \textbf{F1} & \textbf{mIoU} \\ \cline{2-8}
      & 2 & 0.8032 & 0.8126 & 0.7994 & 0.8474 & 0.8231 & 0.8365 \\ 
      & 4 & 0.8123 & 0.8184 & 0.8146 & \underline{0.8523} & 0.8330 & 0.8427 \\ \hline
    \multirow{2}[2]{*}{\scriptsize\begin{sideways}PaSna\end{sideways}} & 2 & 0.8035 & 0.8124 & 0.8062 & 0.8458 & 0.8258 & 0.8369 \\ 
      & 4 & 0.8102 & 0.8162 & 0.8219 & 0.8365 & 0.8291 & 0.8408 \\ \hline
    \multirow{2}[1]{*}{\scriptsize\begin{sideways}Diag\end{sideways}} & 2 & 0.8080 & 0.8166 & 0.8058 & 0.8496 & 0.8271 & 0.8408 \\ 
      & 4 & 0.8091 & 0.8148 & 0.8225 & 0.8417 & 0.8320  & 0.8410  \\ \hline
    \multirow{2}[1]{*}{\scriptsize\begin{sideways}DigSna\end{sideways}} & 2 & 0.8094 & 0.8162 & 0.8185 & 0.8470 & 0.8325 & 0.8413  \\ 
      & 4 & \underline{0.8153}  & \underline{0.8215}  & \underline{0.8237}  & 0.8497  & \underline{0.8365}  & \underline{0.8451} \\ \hline
    \multirow{2}[1]{*}{\scriptsize\begin{sideways}\textbf{SASS}\end{sideways}} & 2 & 0.8130  & 0.8192  & 0.8196  & 0.8478  & 0.8335  & 0.8430  \\ 
      & 4 & \textbf{0.8204} & \textbf{0.8255} & \textbf{0.8241} & \textbf{0.8545} & \textbf{0.8390} & \textbf{0.8479} \\ 
    \bottomrule
    \end{tabular}%
    \caption{Ablation study on the number of paths in different scanning strategies. N represents the number of paths. For two-path scans, SASS uses the first parallel snake and diagonal snake scans, while other methods use the first two paths. Best results are in bold, and second-best results are underlined.}
    \label{tab:scan_num}

\end{table}%

\begin{table}[!ht]
    \centering
    \begin{tabular}{p{0.8cm} | >
    {\centering\arraybackslash}p{0.76cm} >{\centering\arraybackslash}p{0.76cm} >{\centering\arraybackslash}p{0.76cm} >{\centering\arraybackslash}p{0.76cm} >{\centering\arraybackslash}p{0.76cm} >{\centering\arraybackslash}p{0.76cm}}
    \toprule
         $\alpha$ : $\beta$ & ODS & OIS & P & R & F1 & mIoU \\ \hline
        BCE  & 0.8099  & 0.8151  & 0.8207  & 0.8457  & 0.8330  & 0.8414  \\ 
        Dice & 0.8022  & 0.8072  & 0.8038  & 0.8430  & 0.8231  & 0.8358  \\ 
         5 : 1 & 0.8125  & 0.8168  & 0.8207  & 0.8432  & 0.8319  & 0.8428  \\ 
         4 : 1 & 0.8144  & 0.8184  & 0.8217  & 0.8442  & 0.8328  & 0.8437  \\ 
         3 : 1 & 0.8180  & 0.8229  & \textbf{0.8293}  & 0.8436  & 0.8364  & 0.8463  \\ 
         2 : 1 & 0.8098  & 0.8152  & 0.8204  & 0.8392  & 0.8297  & 0.8408  \\ 
         1 : 1 & 0.8123  & 0.8184  & 0.8141  & 0.8507  & 0.8320  & 0.8423  \\ 
         1 : 2 & \underline{0.8152}  & \underline{0.8214}  & 0.8210  & 0.8484  & \underline{0.8345}  & \underline{0.8443}  \\ 
         1 : 3 & 0.8109  & 0.8163  & 0.8226  & 0.8396  & 0.8310  & 0.8418  \\ 
         1 : 4 & 0.8133  & 0.8185  & 0.8163  & \underline{0.8515}  & 0.8336  & 0.8433  \\ 
         \textbf{1 : 5} & \textbf{0.8204}  & \textbf{0.8255}  & \underline{0.8241}  & \textbf{0.8545}  & \textbf{0.8390}  & \textbf{0.8479}  \\ 

    \bottomrule
    \end{tabular}
    \caption{Sensitivity analysis experiments with different $\alpha$ and $\beta$ ratios. Best results are in bold, and second-best results are underlined.}
    \label{tab:loss_sen}
\end{table}

\section{Details of Objective Function and Analysis}
\label{sec:Loss_Details}

The calculation formulas for BCE \cite{li2024rediscovering} loss and Dice \cite{sudre2017generalised} loss are as follows:

\begin{equation}
  L_{Dice} = 1 - \frac{2 \sum_{j=1}^M p_j \hat{p}_j + \epsilon}{\sum_{j=1}^M p_j + \sum_{j=1}^M \hat{p}_j + \epsilon}
\end{equation}

\begin{equation}
L_{BCE} = -\frac{1}{N} \left[ p_j \log(\hat{p}_j) + (1 - p_j) \log(1 - \hat{p}_j) \right]
\end{equation}
where \( M \) denotes the number of samples, \( p_j \) is the ground truth label for the \( j \)-th sample, \( \hat{p}_j \) is the predicted probability for the \( j \)-th sample, \( \epsilon \) is a small constant.

In equation \ref{eq:loss}, the ratio of \( \alpha \) to \( \beta \) is set to 1:5. This is the optimal ratio of $\alpha$ and $\beta$ selected after experimenting with various hyperparameter settings on multi-scenario dataset. As listed in Table \ref{tab:loss_sen}, setting the $\alpha$ to $\beta$ ratio at 1:5 yields the best performance, with improvements of 0.65\% in F1 and 0.55\% in mIoU over the 1:2 ratio. This suggests that balancing Dice and BCE loss at a 1:5 ratio helps the model better distinguish background pixels from the few crack region pixels, thereby enhancing performance.

% \section{Additional Comparison Experiments}
% \label{sec:Other_Comparison}
% To further evaluate the robustness and stability of our method, we conducted experiments on two additional classic public datasets, alongside the four mentioned in the main text.

% \noindent \textbf{CFD} \cite{shi2016automatic}\textbf{.} This dataset contains crack images on asphalt pavement with noise from shadows, oil stains, and water stains. It includes 118 manually annotated images, providing a challenging environment for crack segmentation models.

% \noindent \textbf{GAPS} \cite{eisenbach2017get}\textbf{.} Collected under dry and warm conditions, this dataset includes cracks, potholes, embedded patches, layered patches, and cracked joints, with a total of 509 RGB images for fair comparison across methods.

% Similarly, we divided these two datasets into training, testing, and validation sets with a 7:1:2 ratio. The experimental settings are identical to those in Subsection.

\begin{figure*}[htbp]
  \centering
  \includegraphics[width=\textwidth]{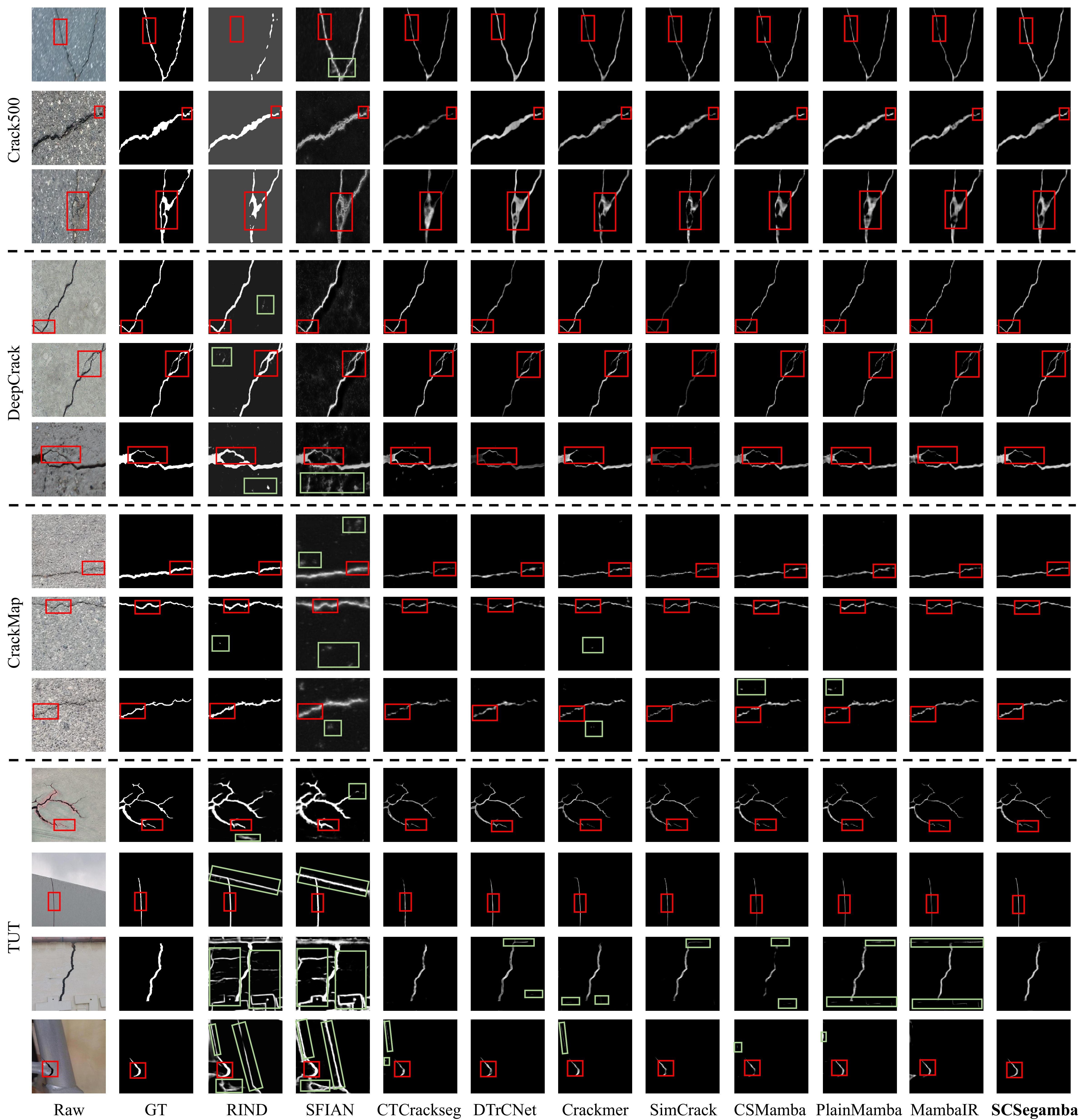}
  \caption{Visual comparison with 9 SOTA methods across four public datasets. Red boxes highlight critical details, and green boxes mark misidentified regions.}
  \label{fig:Com_4_datasets}
\end{figure*}

% As listed in Table \ref{tab:com_add}, our proposed SCSegamba achieved the best performance on both datasets. On the CFD \ref{fig:Com_CFD_GAPS} dataset, the F1 and mIoU scores were 4.61\% and 2.81\% higher than the next best method, respectively. Similarly, on the GAPS \cite{eisenbach2017get} dataset, our method led by 1.57\% in F1 and 0.98\% in mIoU. As shown in Figure \ref{fig:Com_CFD_GAPS}, the crack patterns in the CFD and GAPS datasets are particularly slender. Compared to other methods, SCSegamba better suppresses background interference and accurately captures fine crack details. This is due to SAVSS’s comprehensive perception of crack shapes and textures, combined with MFS’s effective modeling of feature relationships. For CNN-based methods like RIND \cite{pu2021rindnet} and SFIAN \cite{cheng2023selective}, inadequate background noise suppression results from their limited ability to effectively capture contextual information of crack pixels. In contrast, Transformer and Mamba-based models demonstrate strong noise suppression but tend to produce discontinuous segmentation in fine crack details due to suboptimal capture of pixel-level textures.

\begin{table*}[!ht]
    \centering
    \begin{tabular}{cccccccccc}
    \hline
        Layer Num & ODS & OIS & P & R & F1 & mIoU & Params ↓ & FLOPs ↓ & Model Size ↓ \\ \hline
        2 & 0.8102  & 0.8165  & 0.8181  & 0.8420  & 0.8299  & 0.8413  & \textbf{1.56M} & \textbf{12.26G} & \textbf{20MB} \\ 
        4 & \textbf{0.8204}  & \textbf{0.8255}  & \textbf{0.8241}  & \underline{0.8545}  & \textbf{0.8390}  & \textbf{0.8479}  & \underline{2.80M} & \underline{18.16G} & \underline{37MB} \\ 
        8 & \underline{0.8174}  & \underline{0.8222}  & 0.8199  & \textbf{0.8579}  & \underline{0.8387}  & \underline{0.8461}  & 5.23M & 29.27G & 68MB \\ 
        16 & 0.8126  & 0.8187  & \underline{0.8226}  & 0.8475  & 0.8349  & 0.8430  & 10.08M & 51.51G & 127MB \\ 
        32 & 0.5203  & 0.5365  & 0.5830  & 0.5680  & 0.5754  & 0.6785  & 19.79M & 95.97G & 247MB \\ \hline
    \end{tabular}
    \caption{Experiments with different numbers of SAVSS layers. Best results are in bold, and second-best results are underlined.}
    \label{exp:layer_num}
\end{table*}

\begin{table*}[!ht]
    \centering
    \begin{tabular}{cccccccccc}
    \hline
        Patch Size & ODS & OIS & P & R & F1 & mIoU & Params ↓ & FLOPs ↓ & Model Size ↓ \\ \hline
        4 & \underline{0.8053}  & \underline{0.8128}  & \underline{0.8146}  & \underline{0.8443}  & \underline{0.8294}  & \underline{0.8381}  & \textbf{2.61M} & 51.81G & \textbf{34MB} \\ 
        8 & \textbf{0.8204}  & \textbf{0.8255}  & \textbf{0.8241}  & \textbf{0.8545}  & \textbf{0.8390}  & \textbf{0.8479}  & \underline{2.80M} & 18.16G & \underline{37MB} \\ 
        16 & 0.7910  & 0.7937  & 0.8126  & 0.8141  & 0.8133  & 0.8272  & 3.59M & \underline{9.74G} & 45MB \\ 
        32 & 0.7318  & 0.7364  & 0.7535  & 0.7576  & 0.7555  & 0.7879  & 6.74M & \textbf{7.64G} & 82MB \\ \hline
    \end{tabular}
    \caption{Experiments with different patch sizes. Best results are in bold, and second-best results are underlined.}
    \label{exp:patch_size}
\end{table*}

\begin{table*}[!ht]
    \centering
    \begin{tabular}{cccccccccc}
    \hline
        Methods & ODS & OIS & P & R & F1 & mIoU & Params ↓ & FLOPs ↓ & Model Size ↓ \\ \hline
        MambaIR \cite{guo2024mambair} & \underline{0.7869}  & \underline{0.7956}  & \underline{0.7714}  & 0.8445  & \underline{0.8071}  & \underline{0.8240}  & 3.57M & 19.71G & \underline{29MB} \\ 
        CSMamba \cite{liu2024cmunet} & 0.7140  & 0.7201  & 0.6934  & 0.8171  & 0.7503  & 0.7773  & 12.68M & \underline{15.44G} & 84MB \\ 
        PlainMamba \cite{yang2024plainmamba} & 0.7787  & 0.7896  & 0.7617  & \underline{0.8531}  & 0.8064  & 0.8201  & \textbf{2.20M} & \textbf{14.09G} & \textbf{18MB} \\ 
        SCSegamba (\textbf{Ours}) & \textbf{0.8204}  & \textbf{0.8255}  & \textbf{0.8241}  & \textbf{0.8545}  & \textbf{0.8390}  & \textbf{0.8479}  & \underline{2.80M} & 18.16G & 37MB \\ \hline
    \end{tabular}
    \caption{Comparison experiments of different Mamba-based methods using 4 VSS layers. Best results are in bold, and second-best results are underlined.}
    \label{exp:cmp_layer_4}
\end{table*}
\vspace{0.5cm}

\section{Visualisation Comparisons}
\label{sec:Visualisation_Comparisons}

\begin{figure*}[htbp]
  \centering
  \includegraphics[width=0.9\textwidth]{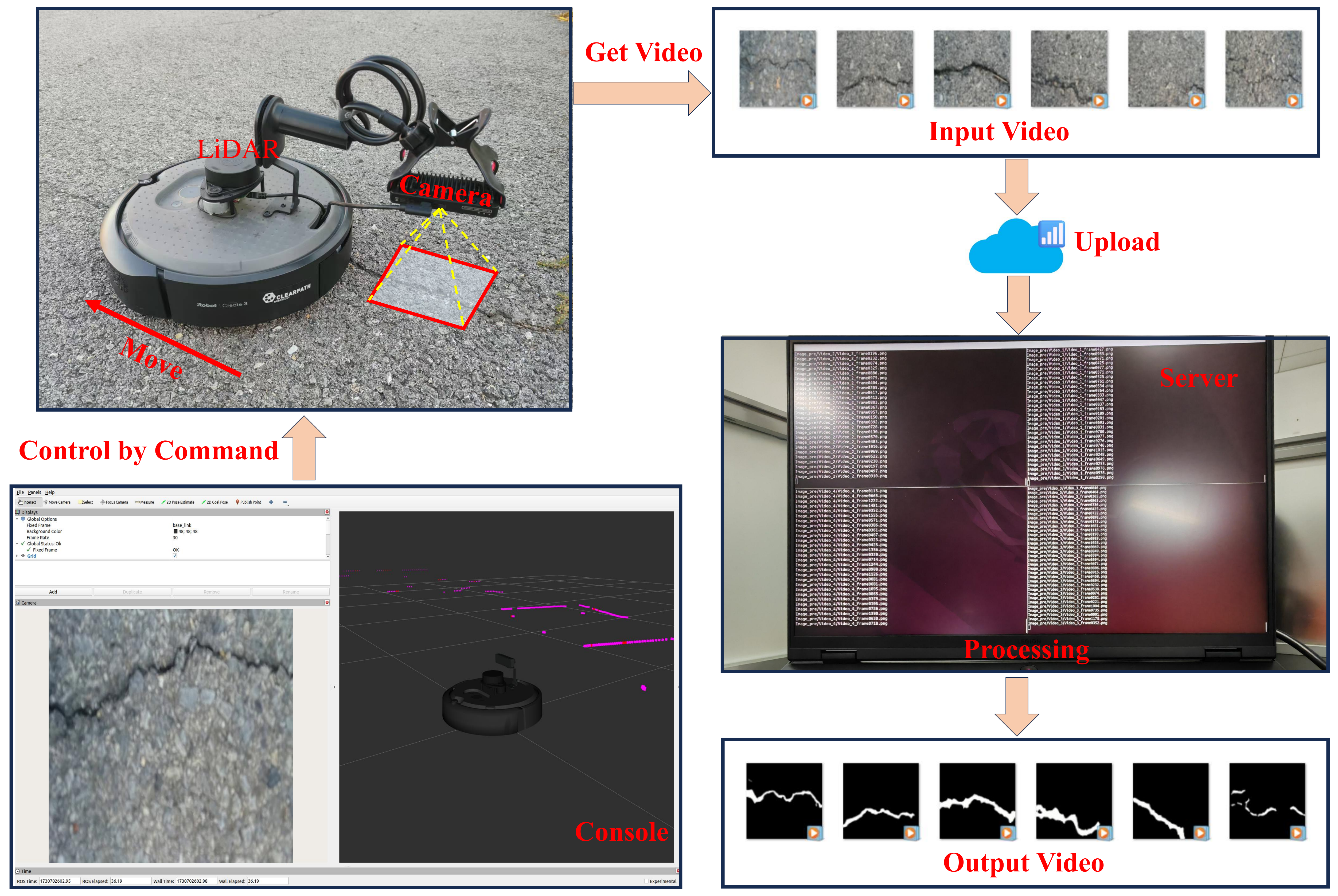}
  \caption{Schematic of real-world deployment. The intelligent vehicle is placed on an outdoor road surface, and we use the server terminal to remotely control it. The vehicle transmits the video data in real-time to the server, where it is processed to obtain the final output.}
  \label{fig:real_app}
  \vspace{-0.3cm}
\end{figure*}

To visually demonstrate the advantages of SCSegamba, we present detailed visual results in Figure \ref{fig:Com_4_datasets}. For the Crack500 \cite{yang2019feature}, DeepCrack \cite{liu2019deepcrack}, and CrackMap \cite{katsamenis2023few} datasets, which primarily include bitumen, concrete, and brick scenarios with minimal background noise and a range of crack thicknesses, our method consistently achieves accurate segmentation, even capturing intricate fine cracks. This is attributed to GBC’s strong capability in capturing crack morphology. In contrast, other methods show weaker performance in continuity and fine segmentation, resulting in discontinuities and expanded segmentation areas that do not align with actual crack images.

For the TUT \cite{liu2024staircase} dataset, which includes diverse scenarios and significant background noise, our method excels at suppressing interference. For instance, in images of cracks on generator blades and steel pipes, it effectively minimizes irrelevant noise and provides precise crack segmentation. This performance is largely attributed to SAVSS's accurate capture of crack topologies. In contrast, CNN-based methods like RIND \cite{pu2021rindnet} and SFIAN \cite{cheng2023selective} struggle to distinguish background noise from crack regions, highlighting their limitations in contextual dependency capture. Other Transformer and Mamba-based methods also fall short in segmentation continuity and detail handling compared to our approach.

\section{Additional Analysis}
\label{sec:Additional_Ablation}

To provide a thorough demonstration of the necessity of each component in our proposed SCSegamba, we conducted a more extensive analysis experiment.

\noindent \textbf{Comparison with different numbers of SAVSS layers.}
In our SCSegamba, we used 4 layers of SAVSS blocks to balance performance and computational requirements. As listed in Table \ref{exp:layer_num}, 4 layers achieved optimal results, with F1 and mIoU scores 0.036\% and 0.21\% higher than with 8 layers, while reducing parameters by 2.43M, computation by 11.11G, and model size by 31MB. Although using only 2 layers minimized resource demands, with 1.56M parameters, performance decreased. Conversely, using 32 layers increased resource use and reduced performance due to redundant features, which impacted generalization. Thus, 4 SAVSS layers strike an effective balance between performance and resource efficiency, making it ideal for practical applications.

\begin{figure}[htbp]
  \centering
  \includegraphics[width=0.48\textwidth]{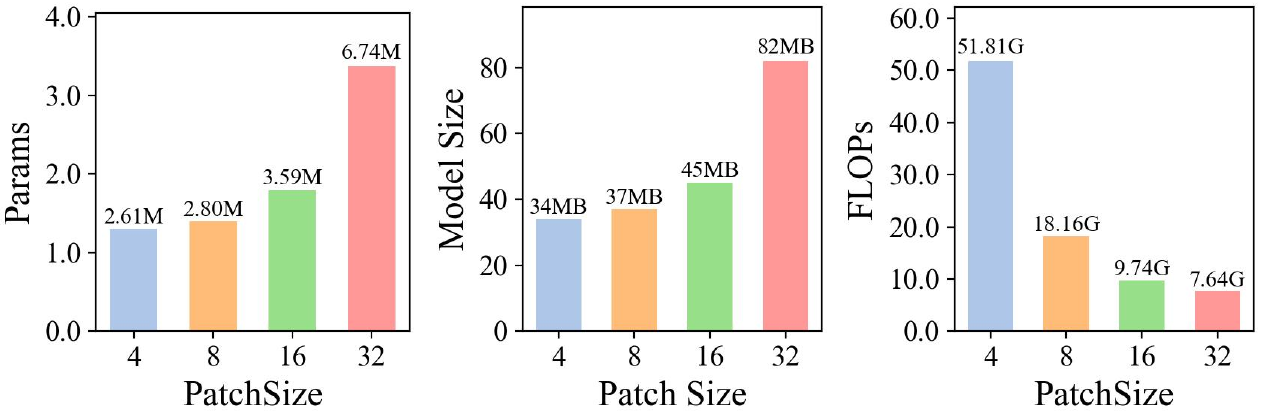}
  \caption{Comparison of computing resources required for different Patch Size}
  \label{fig:patch_ana}
\end{figure}

\noindent \textbf{Comparison with different Patch Size.}
In our SAVSS, we set the Patch Size to 8 during Patch Embedding. To verify its effectiveness, we conducted experiments with various Patch Sizes. As listed in Table \ref{exp:patch_size}, a Patch Size of 8 yields the best performance, with F1 and mIoU scores 1.16\% and 1.17\% higher than a Patch Size of 4. Although a smaller Patch Size of 4 reduces parameters and model size, it limits the receptive field and hinders the effective capture of longer textures, impacting segmentation. As shown in Figure \ref{fig:patch_ana}, as the Patch Size increases, parameter count and model size decrease, but the computational load per patch rises, affecting efficiency. At a Patch Size of 32, performance drops significantly due to reduced fine-grained detail capture and sensitivity to contextual variations. Thus, a Patch Size of 8 balances detail accuracy and generalization while maintaining model efficiency.

\noindent \textbf{Comparison under the same number of VSS layers.}
In Subsection \ref{subsec:com_sota}, we compare SCSegamba with other SOTA methods, using default VSS layer settings for Mamba-based models like MambaIR \cite{guo2024mambair}, CSMamba \cite{liu2024cmunet}, and PlainMamba \cite{yang2024plainmamba}. To examine complexity and performance under uniform VSS layer counts, we set all Mamba-based models to 4 VSS layers and conducted comparisons. As listed in Table \ref{tab:Complex_Com} and \ref{exp:cmp_layer_4}, although computational requirements for MambaIR, CSMamba, and PlainMamba decrease, their performance drops significantly. For example, CSMamba's F1 and mIoU scores drop to 0.7503 and 0.7773. While PlainMamba with 4 layers achieves reductions of 0.60M in parameters, 4.07G in FLOPs, and 19MB in model size, SCSegamba surpasses it by 4.04\% in F1 and 3.39\% in mIoU. Thus, with 4 SAVSS layers, SCSegamba balances performance and efficiency, capturing crack morphology and texture for high-quality segmentation.

\section{Real-world Deployment Applications}
\label{sec:Real_world}

To validate the effectiveness of our proposed SCSegamba in real-world applications, we conducted a practical deployment and compared its real-world performance with other SOTA methods. Specifically, our experimental system consists of two main components: the intelligent vehicle and the server. The intelligent vehicle used is a Turtlebot4 Lite driven by a Raspberry Pi 4, equipped with a LiDAR and a camera. The camera model is OAK-D-Pro, fitted with an OV9282 image sensor capable of capturing high-quality crack images. The server is a laptop equipped with a Core i9-13900 CPU running Ubuntu 22.04. The intelligent vehicle and server communicate via the internet. This setup simulates resource-limited equipment to evaluate the performance of our SCSegamba in real-world deployment scenarios.

\begin{table}[!ht]
    \centering
    \begin{tabular}{cc}
    \hline
        Methods & Inf Time↓ \\ \hline
        RIND \cite{pu2021rindnet} & 0.0909s \\ 
        SFIAN \cite{cheng2023selective} & 0.0286s \\ 
        CTCrackseg \cite{tao2023convolutional} & 0.0357s \\ 
        DTrCNet \cite{xiang2023crack} & 0.0213s \\ 
        Crackmer \cite{wang2024dual} & 0.0323s \\ 
        SimCrack \cite{jaziri2024designing} & 0.0345s \\ 
        CSMamba \cite{liu2024cmunet} & 0.0625s \\ 
        PlainMamba \cite{yang2024plainmamba} & 0.1667s \\ 
        MambaIR \cite{guo2024mambair} & 0.0400s \\ 
        SCSegamba (\textbf{Ours}) & 0.0313s \\ \hline
    \end{tabular}
    \caption{Comparison of inference time with other SOTA methods on resource-constrained server.}
    \label{tab:fps}
\end{table}

As shown in Figure \ref{fig:real_app}, in the real-world deployment process, the intelligent vehicle was placed on an outdoor road surface filled with cracks. We remotely controlled the vehicle from the server terminal, directing it to move forward in a straight line at a speed of 0.15 m/s. The camera captured video at a frame rate of 30 frames per second. The vehicle transmitted the recorded video data to the server in real-time via the network. To accelerate data transmission from the vehicle to the server, we set the recording resolution to 512 × 512. Upon receiving the video data, the server first segmented it into frames, then fed each frame into the pre-trained SCSegamba model, which was trained on all datasets, for inference. After segmentation, the server recombined the processed frames into a video, yielding the final output. This setup simulates real-time crack segmentation in an real-world production process.

Additionally, we deployed the weight files of other SOTA methods on the server for comparison. As listed in Table \ref{tab:fps}, our SCSegamba achieved an inference speed of 0.0313 seconds per frame on the resource-constrained server, outperforming most other methods. This demonstrates that our method has excellent real-time performance, making it suitable for real-time segmentation of cracks in video data.

\begin{figure*}[htbp]
  \centering
  \includegraphics[width=0.8\textwidth]{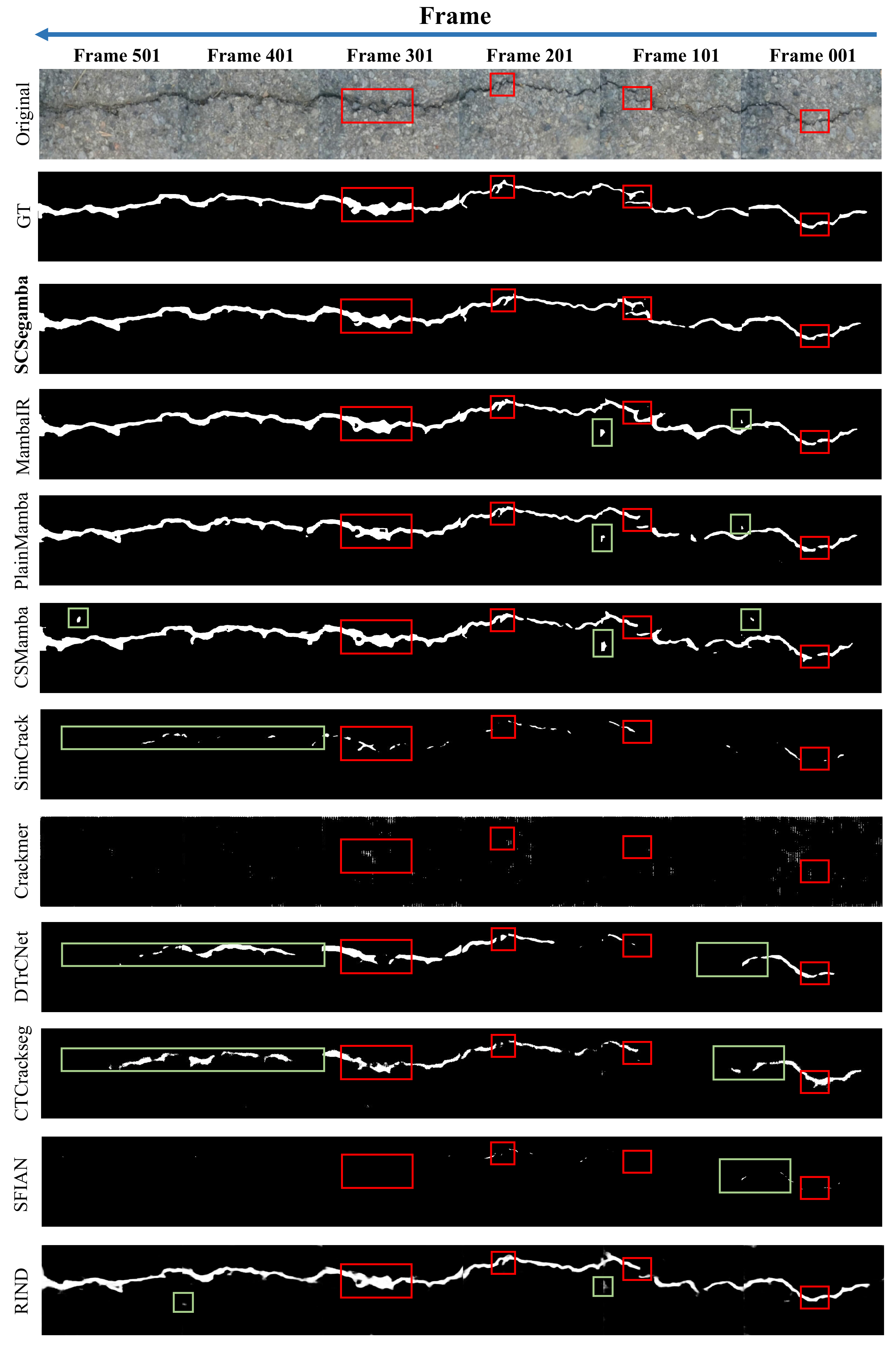}
  \caption{Visualisation comparison on video data keyframes. The interval between keyframes is 100 frames in order to ensure continuity of observation. Red boxes highlight critical details, and green boxes mark misidentified regions.}
  \label{fig:real_app_results}
\end{figure*}

As shown in Figure \ref{fig:real_app_results}, compared to other SOTA methods, our SCSegamba better suppresses irrelevant noise in video data and generates continuous crack region segmentation maps. For instance, although SSM-based methods like PlainMamba \cite{yang2024plainmamba}, MambaIR \cite{guo2024mambair}, and CSMamba \cite{liu2024cmunet} achieve continuous segmentation, they tend to produce false positives in some irrelevant noise spots. Additionally, while CNN and Transformer-based methods achieve high metrics and performance on datasets with faster inference speed, their performance on video data is suboptimal, often showing discontinuous segmentation and poor background suppression. For example, cracks segmented by DTrCNet \cite{xiang2023crack} and CTCrackSeg \cite{tao2023convolutional} exhibit significant discontinuities, and Crackmer \cite{wang2024dual} struggles to distinguish between crack and background regions. Based on the above real-world deployment results, our SCSegamba produces high-quality segmentation results on crack video data with low parameters and computational resources, making it more suitable for deployment on resource-constrained devices and demonstrating its strong performance in practical production scenarios.

\begin{algorithm}[htbp]
\caption{SASS execution process}
\label{alg:sass}
\begin{algorithmic}[1]
\STATE \textbf{Input:} Patch matrix dimensions $H$, $W$
\STATE \textbf{Output:} $O = (o1, o2, o3, o4)$, $O\_inverse = (o1\_inverse, o2\_inverse, o3\_inverse, o4\_inverse)$, $D = (d1, d2, d3, d4)$

\STATE \textbf{Initialize:} $L = H \times W$

\STATE Initialize $(i, j) \gets (0, 0)$ for $o1$, $(H - 1, W - 1)$ if $H$ is odd else $(H - 1, 0)$ for $o2$
\STATE $i_d \gets down$, $j_d \gets left$ if $H$ is odd else $right$

\WHILE{$j < W$ or $i \geq 0$}
    \STATE $idx \gets i \times W + j$, append $idx$ to $o1$, set $o1\_inverse[idx]$
    \IF{$i_d = down$ and $i < H - 1$}
        \STATE $i \gets i + 1$, add $down$ to $d1$
    \ELSE
        \STATE $j \gets j + 1$, $i_d \gets up$ if $i = H - 1$ else $down$, add $right$ to $d1$
    \ENDIF
    \STATE $idx \gets i \times W + j$, append $idx$ to $o2$, set $o2\_inverse[idx]$
    \IF{$j_d = right$ and $j < W - 1$}
        \STATE $j \gets j + 1$, add $right$ to $d2$
    \ELSE
        \STATE $i \gets i - 1$, $j_d \gets left$ if $j = W - 1$ else $right$, add $up$ to $d2$
    \ENDIF
\ENDWHILE
    
\STATE $d1 \gets [d_{start}] + d1[:-1]$, $d2 \gets [d_{start}] + d2[:-1]$

\FOR{$\text{diag} \gets 0$ \textbf{to} $H + W - 2$}
    \STATE $direction \gets right$ if $\text{diag}$ is even else $down$
    \FOR{$k \gets 0$ \textbf{to} $\min(\text{diag} + 1, H, W) - 1$}
        \STATE $i, j \gets (\text{diag} - k, k)$ if $\text{diag}$ is even else $(k, \text{diag} - k)$
        \IF{$j < W$}
            \STATE $idx \gets i \times W + j$
            \STATE Append $idx$ to $o3$, set $o3\_inverse[idx]$, add $direction$ to $d3$
        \ENDIF
        \STATE $i, j \gets (\text{diag} - k, W - k - 1)$ if $\text{diag}$ is even else $(k, W - \text{diag} + k - 1)$
        \IF{$j < W$}
            \STATE $idx \gets i \times W + j$
            \STATE Append $idx$ to $o4$, set $o4\_inverse[idx]$, add $direction$ to $d4$
        \ENDIF
    \ENDFOR
\ENDFOR
\STATE $d3 \gets [d_{start}] + d3[:-1]$, $d4 \gets [d_{start}] + d4[:-1]$

\STATE \textbf{Return:} $O$, $O\_inverse$, $D$

% \STATE \textbf{Return:} $O = (o1, o2, o3, o4)$, $O\_inverse = (o1\_inverse, o2\_inverse, o3\_inverse, o4\_inverse)$, $D = (d1, d2, d3, d4)$

\end{algorithmic}
\end{algorithm}
\vspace{2.0cm}

% \section{Rationale}
% \label{sec:rationale}
% % 
% Having the supplementary compiled together with the main paper means that:
% % 
% \begin{itemize}
% \item The supplementary can back-reference sections of the main paper, for example, we can refer to \cref{sec:intro};
% \item The main paper can forward reference sub-sections within the supplementary explicitly (e.g. referring to a particular experiment); 
% \item When submitted to arXiv, the supplementary will already included at the end of the paper.
% \end{itemize}
% % 
% To split the supplementary pages from the main paper, you can use \href{https://support.apple.com/en-ca/guide/preview/prvw11793/mac#:~:text=Delete%20a%20page%20from%20a,or%20choose%20Edit%20%3E%20Delete).}{Preview (on macOS)}, \href{https://www.adobe.com/acrobat/how-to/delete-pages-from-pdf.html#:~:text=Choose%20%E2%80%9CTools%E2%80%9D%20%3E%20%E2%80%9COrganize,or%20pages%20from%20the%20file.}{Adobe Acrobat} (on all OSs), as well as \href{https://superuser.com/questions/517986/is-it-possible-to-delete-some-pages-of-a-pdf-document}{command line tools}.